%% file: main_tcyb.tex
\documentclass[journal]{IEEEtran}

\usepackage[utf8]{inputenc}
\usepackage[T1]{fontenc}
\usepackage{cite}
\usepackage{url}
\usepackage{booktabs}
\usepackage{nicefrac}
\usepackage{microtype}
\usepackage{xcolor}
\usepackage{graphicx}
\graphicspath{{figures/}}
\usepackage{subcaption}
\usepackage[ruled,vlined,linesnumbered]{algorithm2e}
\usepackage{placeins}
\usepackage{multirow}
\usepackage{amsmath,amssymb,amsfonts}
\usepackage{amsthm}
\usepackage[hidelinks]{hyperref}

\newcommand{\pms}[1]{{\scriptsize$\pm$#1}}
\DontPrintSemicolon
\SetKwInput{KwIn}{Input}
\SetKwInput{KwOut}{Output}
\newtheorem{definition}{Definition}

\title{Edge-Gradient Augmentation under Reinforcement Learning for Temporal Graph}
\author{Yongqing~Zhou, Jiawei~Chen, Wenjun~Wang, Xinyu~Wang, Wei~Yu, and Qiyao~Peng%
\thanks{\raggedright Yongqing Zhou, Jiawei Chen, Wenjun Wang, Xinyu Wang, and Qiyao Peng are with Tianjin University, Tianjin, China
(\mbox{e-mail:} zyq\_901@tju.edu.cn; ppw@tju.edu.cn; wjwang@tju.edu.cn; xinxinxin@tju.edu.cn; qypeng@tju.edu.cn).
Wei Yu is with Zhejiang Yuexiu University, Shaoxing, China (\mbox{e-mail:} weiyu@tju.edu.cn).
\textit{Corresponding author: Qiyao~Peng.}}}
\hypersetup{pdfauthor={Yongqing Zhou, Jiawei Chen, Wenjun Wang, Xinyu Wang, Wei Yu, Qiyao Peng}}

\begin{document}

\maketitle

\begin{abstract}
Temporal link prediction is critical for understanding dynamic interactions in evolving networks, informing applications ranging from social network analysis to recommendation systems. Existing methods typically rely on raw temporal graph data, often neglecting explicit edge-change information that can significantly improve predictive performance. To address this gap, we propose edge-gradient reinforced learning (EG-RL), a framework that explicitly encodes edge-gradient events (edge births and deaths) through dedicated node and link encoding modules. EG-RL uses a policy-driven reinforcement-learning mechanism to dynamically select and augment the most informative edge snapshots, guided by critic-based feedback to optimize prediction accuracy. The model comprises distinct modules for node encoding, link encoding, token mixing, and channel mixing, integrated through global pooling and followed by a two-layer multilayer perceptron classifier. Extensive experiments on multiple dynamic graph datasets demonstrate substantial gains in area under the receiver operating characteristic curve and average precision. Ablation studies further confirm the necessity of adaptive edge-gradient augmentation and validate the effectiveness of the proposed encoding and reinforcement-learning mechanisms. To encourage further exploration, we make our codebase and processed datasets publicly available at https://anonymous.4open.science/r/EG-RL-671D.
\end{abstract}

\begin{IEEEkeywords}
dynamic graphs, graph representation learning, reinforcement learning, temporal link prediction
\end{IEEEkeywords}

\section{Introduction}
Temporal link prediction forecasts future interactions from time-ordered observations, supporting recommendation, communication analysis, transaction monitoring, and knowledge-graph forecasting \cite{kazemi2020representation,wu2021temporal}. Beyond interaction recency, a predictor must distinguish persistent activity, isolated encounters, and disappearance followed by reappearance. These patterns vary across domains, motivating representations that preserve temporal and structural context without assigning equal relevance to every historical observation.

Existing temporal graph methods model structural evolution through recurrent and snapshot updates \cite{pareja2020evolvegcn,kumar2019predicting,sankar2020dysat,you2022roland}, temporal attention and event-driven memory \cite{xuinductive,tgn}, and asynchronous propagation and motif-aware walks \cite{wang2021apan,jin2022neural}. Formation and dissolution are also considered in continuous-time temporal knowledge graphs \cite{han2021temporal}. We investigate a complementary question: can an explicit, interpretable encoding of observed edge transitions provide a useful inductive bias, and can selecting among the resulting augmented observations improve prediction within a simple temporal encoder?

\begin{figure*}[t]
    \centering
    \begin{subfigure}[b]{0.44\linewidth}
        \centering
        \includegraphics[width=\linewidth]{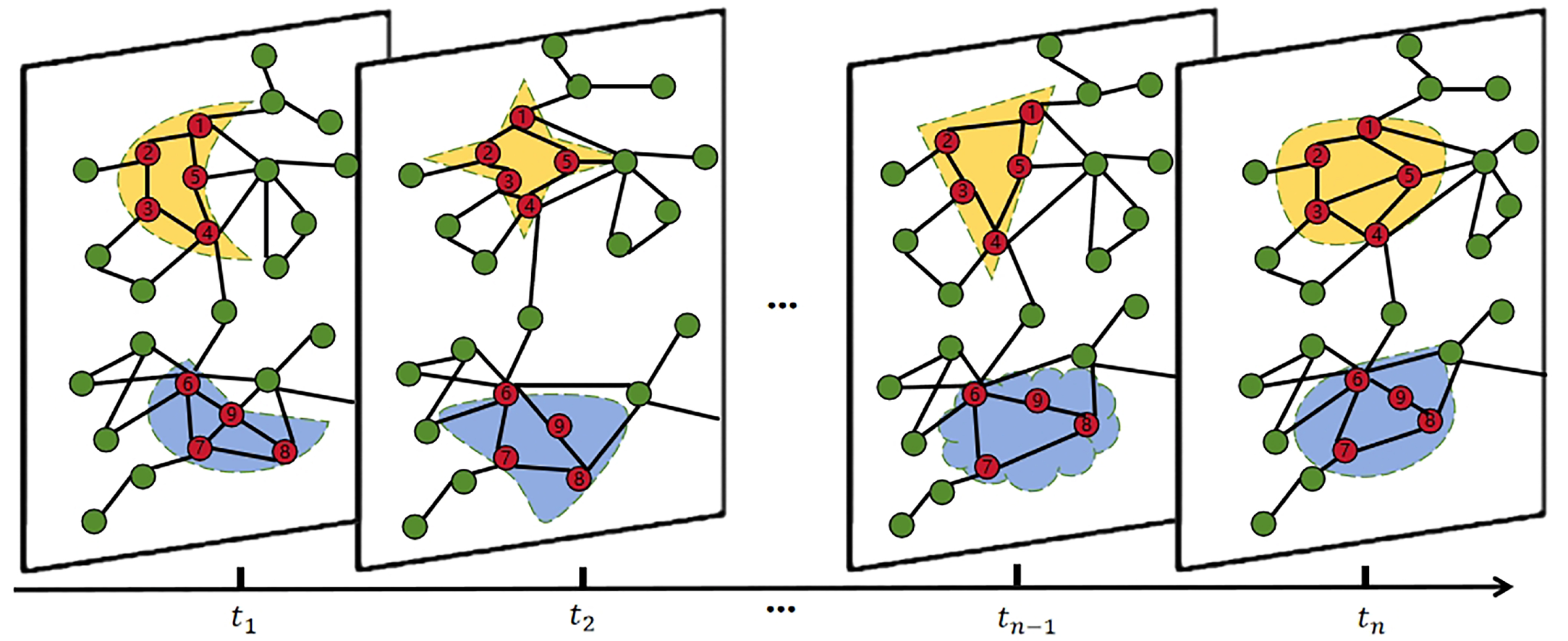}
        \caption{Original Temporal Graph}
        \label{fig:A}
    \end{subfigure}
    \hspace{0.015\linewidth}
    \raisebox{10ex}{{\Large$\Longrightarrow$}}
    \hspace{0.015\linewidth}
    \begin{subfigure}[b]{0.44\linewidth}
        \centering
        \includegraphics[width=\linewidth]{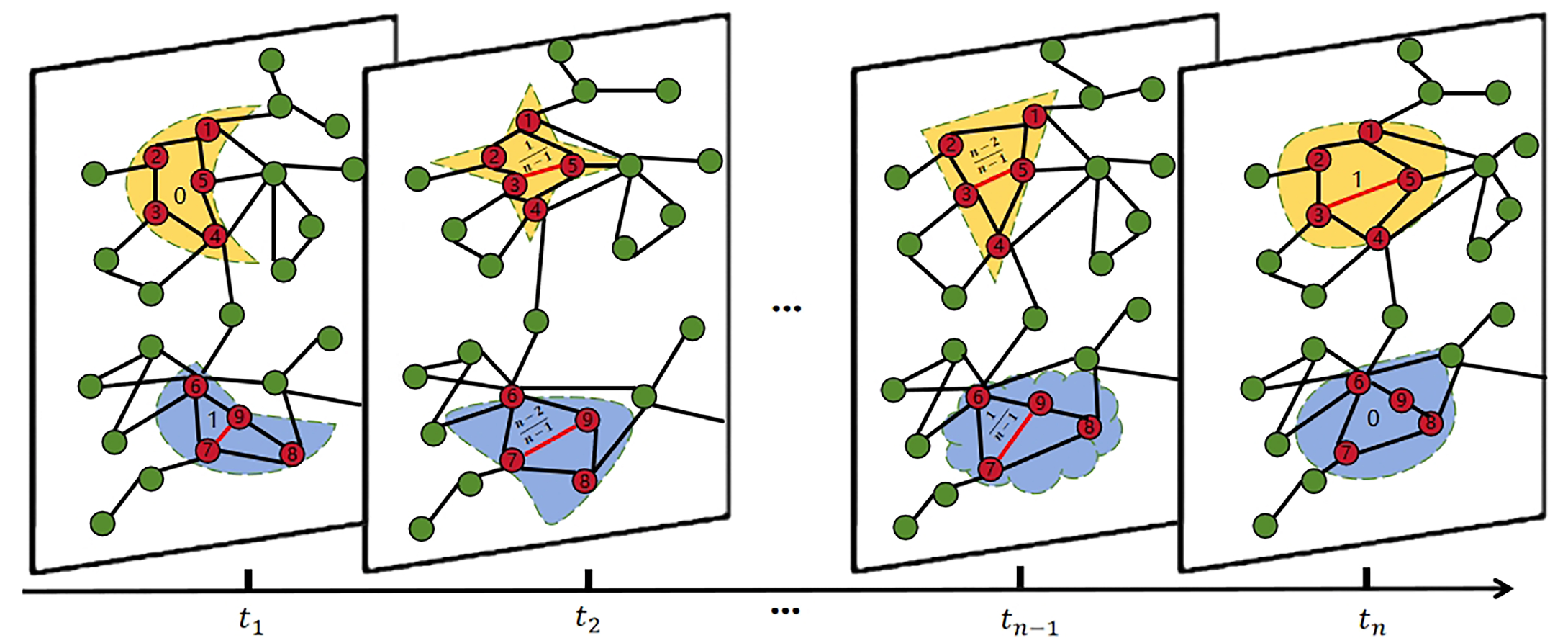}
        \caption{Edge-Gradient Augmentation}
        \label{fig:B}
    \end{subfigure}

    \caption{Illustration of Edge-Gradient Augmentation. 
(a) Original temporal graph snapshots with binary edges at times $t_1,\dots,t_n$. 
(b) Constructed edge-weight profiles between observed transition endpoints. These are modelling signals, not measurements of latent relationship strength; both endpoints must be available before a prediction uses the profile.}

    \label{fig:orig_vs_recon}
\end{figure*}

Pairs with equally recent interactions can have different histories of presence, absence, and reappearance. Edge-Gradient Reinforced Learning (EG-RL) exposes these transitions through constructed edge-weight profiles. Linear interpolation maps a defined $0\!\to\!1$ transition to an increasing profile and a $1\!\to\!0$ transition to a decreasing profile (Fig.~\ref{fig:orig_vs_recon}). This parameter-free transformation reorganizes historical evidence alongside the original timing information; it adds no new observations.

An edge gradient describes changes in recorded presence, not a measured relationship velocity or latent intensity: a slice without a recorded interaction does not establish that a relationship has ended. Its usefulness may depend on the observation regime. Repeated transitions may provide informative context, whereas interpolation is less well supported for isolated events, irregular coverage, or long gaps. Augmented observations are therefore candidates for selection, not uniformly useful inputs.

EG-RL uses a policy-based selector because retention is discrete, candidate utility depends on the jointly retained observations, and validation AP/AUC are not directly differentiable. The policy observes graph and candidate features within a bounded pool; a clipped relative validation-AUC reward supplies feedback, and a critic baseline aims to reduce policy-gradient variance. Credit is shared statistically across a selection episode, not observed separately for each decision. Attention, learned weighting, differentiable gates, and temporal sampling remain plausible alternatives; RL should therefore be compared with these and random or heuristic selection under matched candidates, encoders, and budgets. Held-out test scores are excluded from rewards and checkpoint selection.

Temporal validity requires both interpolation endpoints and their features to be available before the prediction cutoff; constructing profiles from the complete sequence before splitting would risk leakage. Original event clocks and observation granularity must remain explicit. Strong scores under easy negatives can also obscure weaknesses in temporal reasoning \cite{poursafaei2022towards}. Chronological evaluation, historical negative sampling, and new-node inductive evaluation thus provide distinct, non-interchangeable evidence.

We use a GraphMixer-based encoder with node/link features, token/channel mixing, pooling, and an MLP predictor \cite{chen2023graphmixer}, keeping the backbone simple enough to examine augmentation and selection separately. Candidate construction, policy/critic evaluation, and validation feedback nonetheless add overhead, motivating assessment of runtime and peak memory alongside predictive gains. Our eight-dataset study includes baseline comparisons and component ablations.

Our main contributions are summarised as follows:
\begin{itemize}
    \item We formulate \emph{edge-gradient augmentation} as an explicit, parameter-free representation of defined binary-presence transitions, distinguishing constructed weights from latent relationship intensities and specifying the endpoint-availability condition required for causal prediction.
    \item We combine a bounded, discrete policy-based selector with critic feedback and a GraphMixer backbone, separating candidate construction, candidate retention, and temporal encoding so that their individual roles can be examined.
    \item We study temporal link prediction on eight public datasets with baseline comparisons and component ablations, interpreting the evidence in relation to observation semantics and evaluation protocols. The anonymous code reference is provided to support inspection and reproducibility.
\end{itemize}

\section{Related Work}
\subsection{Temporal Link Prediction on Dynamic Graphs}
CTDNE uses time-respecting walks \cite{nguyen2018continuous}, EvolveGCN
recurrently updates GCN parameters \cite{pareja2020evolvegcn}, and DyRep
couples node-state evolution with interaction processes \cite{dyrep}.
TGN maintains event-driven memory \cite{tgn}, whereas TGAT uses functional
time encodings \cite{xuinductive}. Recent directions include community-guided
walks \cite{yu2026ctwalks}, topology-aware transformers \cite{zhang2025topdyg},
and joint Euclidean--hyperbolic representations \cite{sun2025geometric}.
DyGPrompt uses node--time conditional prompts \cite{yu2025dygprompt}; TAG
combines time-aware edge dropping, recent-neighbor sampling, and contrastive
learning \cite{pan2025tag}. TMetaNet guides adaptation across snapshots
using dynamic topological features \cite{li2025tmetanet}. EG-RL instead
constructs and selects explicit edge-transition augmentation.

\subsection{Edge-Intensity and Gradient Modelling}
Time-decayed line graphs weight interactions by temporal proximity
\cite{chanpuriya2023tdlg}. TANGO models continuous-time knowledge-graph
embeddings and explicitly encodes edge formation and dissolution
\cite{han2021temporal}; GN-CDE jointly models node and structural dynamics
using neural controlled differential equations \cite{qin2026gncde}.
TAMI addresses heterogeneous intervals and intermittent interactions
through logarithmic time encoding and link-history aggregation
\cite{yu2025tami}. Thus, prior work does model continuous dynamics and
edge changes. EG-RL differs by constructing parameter-free linear weights
from defined binary-presence transitions, then learning which augmented
observations to retain. These weights are modelling signals, not measured
latent interaction intensities.

Dynamic augmentation must preserve temporal consistency. DyAug separates
rationale and environment representations and augments the latent environment
\cite{zhang2025dyaug}. Conda generates historical-neighbor embeddings using
latent conditional diffusion \cite{tian2024conda}. EG-RL instead interpolates
explicit edge weights and retains candidates with a policy. These approaches
motivate separate tests of candidate construction, selection, and temporal
validity.

\subsection{Reinforcement Learning and Lightweight Mixers}
Reinforcement learning supports graph explanation \cite{shan2021reinforcement},
and continual learning supports streaming graph updates \cite{wang2020streaming}.
EG-RL uses policy--critic feedback for augmentation selection.
GraphMixer demonstrates effective MLP-based temporal prediction
\cite{chen2023graphmixer}; graph transformers and token mixers provide other
encoder designs \cite{hussain2022egt,mialon2021graphit,he2023generalization}.
L-STEP combines learnable spatial--temporal positional encoding with an MLP
\cite{tieu2025lstep}, while CRAFT performs target-aware matching
\cite{yi2025craft}. Frequency-domain alternatives include FreeDyG's
interaction-pattern modelling \cite{tian2024freedyg} and BandRank's band-pass
disentanglement for candidate ranking \cite{li2025bandrank}. GraphSSM extends
state-space modelling to temporal graphs \cite{li2024graphssm}, alongside
scalable temporal relative positional encodings \cite{cheng2025temporalrpe}.
EG-RL focuses on augmentation and retention within a lightweight backbone.

\subsection{Evaluation Protocols and Observation Regimes}
TGB-Seq highlights repeated-edge bias and introduces more demanding
sequential dynamics \cite{yi2025tgbseq}; TGB 2.0 extends evaluation to
large multi-relational temporal graphs \cite{gastinger2024tgb2}.
UTG \cite{huang2024utg} and TGM \cite{chmura2026tgm} bridge event-based
and snapshot-based modelling, emphasizing explicit temporal granularity
and reproducible pipelines. Our eight-dataset evaluation,
however, does not establish performance on these newer benchmark suites
or against all recent architectures.

\smallskip
\noindent
\textbf{Summary.} \;
EG-RL combines explicit transition-based augmentation, reinforcement-guided
selection, and a lightweight mixer; its evaluation must distinguish
observation semantics from predictive gains.
\begin{figure*}[t]
    \centering
    \includegraphics[width=\textwidth]{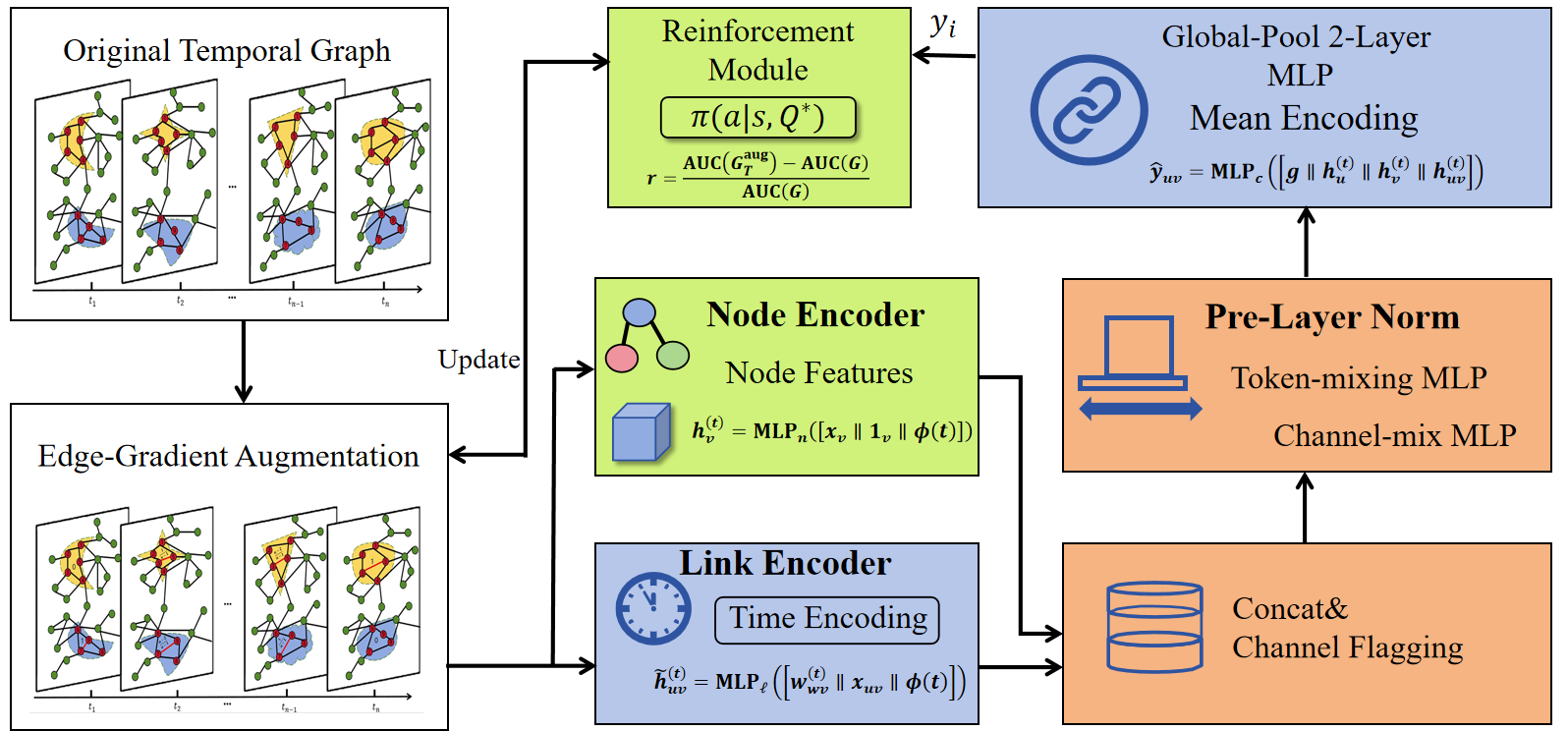}
    \caption{\textbf{Overview of the EG-RL architecture.}
    The \emph{original temporal graph} (top-left) is first converted into an
    \emph{edge-gradient--augmented sequence} (bottom-left) that exposes the pace of interaction changes.
    Every snapshot is fed to a \textbf{node encoder}
    (static features + identity) and a \textbf{link encoder}
    (time encoding + 1-layer MLP with mean pooling).
    The resulting tokens are concatenated and channel-flagged
    ($\alpha_j$), then processed by a stack of
    \textbf{Graph-Mixer blocks} that alternate
    token-mix and channel-mix MLPs under pre-layer normalisation.
    A global mean pool followed by a two-layer MLP
    produces the link-existence logit $y_i$.
    A \textbf{reinforcement module} observes graph embeddings and
    adaptively retains or discards augmented snapshots; its reward
    signal is fed back (\emph{Update}) to refine the augmentation loop.}
    \label{fig:2}
\end{figure*}
\section{Problem Definition and Methodology}
\subsection{Dynamic-Graph Notation and Task}
\label{subsec:notation}
We model a temporal graph as an ordered sequence
$\mathcal{G}_{1:T}=\{G_t=(V,E_t)\}_{t=1}^{T}$,
where $V$ is a fixed node set ($|V|=N$) and
$E_t\subseteq V\times V$ is the edge set observed at time~$t$.
A binary indicator $e_{uv}^{(t)}\in\{0,1\}$ denotes whether edge $(u,v)$ exists at time~$t$.
Given the observed history $\mathcal{G}_{1:T}$, the temporal link prediction task is to estimate the conditional probability that edge $(u,v)$ exists at a future time step $T+\Delta$:
\begin{equation}
  p_{uv}^{(T+\Delta)}
  = \Pr\bigl(e_{uv}^{(T+\Delta)}=1 \mid \mathcal{G}_{1:T}\bigr)
  \label{eq:task_prob}
\end{equation}
where $p_{uv}^{(T+\Delta)}\in[0,1]$ denotes the predicted probability of future edge existence.
In our experiments, we set $\Delta=1$ in Eq.~\eqref{eq:task_prob}.
Evaluation follows prior work using Area Under the ROC Curve (AUC) and Average Precision (AP) on a held-out set balanced by time-matched negatives.

\paragraph{Observation semantics.}
For timestamped event sources,
each distinct original timestamp $\tau_i$ defines one slice $G_i$.
The index $i$ orders observations; it is not a physical time unit, and
$\Delta=1$ denotes the next supplied slice. Multiple records for the same
directed pair within a slice contribute one binary presence. In this
representation, absence means no recorded connection within the observed
scope, not evidence that a latent relationship has permanently ended.
Interpolation is a representation of observed transitions, rather than
a measurement of unobserved physical edge dynamics. For prospective use,
both transition endpoints and their associated features must already be
available before the prediction cutoff; an unobserved endpoint cannot be
used to reconstruct the past input of that prediction.
\subsection{Edge-Gradient Interpolation}
\label{subsec:edgegradient}
Binary snapshots solely record presence or absence, masking \emph{how
fast} interactions strengthen or fade.  We explicitly construct
edge-intensity curves.

\begin{definition}[Edge-gradient]
\label{def:edgegrad}
Let $(t_s,t_e)$ be the first and last timestamps of an edge
transition $0\!\to\!1$ ($1\!\to\!0$ is analogous).  For every
$t_s<t<t_e$ we assign
\begin{equation}
w_{uv}^{(t)}=
\begin{cases}
\dfrac{t-t_s}{t_e-t_s}, & \text{for a }0\!\to\!1\text{ transition},\\[4pt]
1-\dfrac{t-t_s}{t_e-t_s}, & \text{for a }1\!\to\!0\text{ transition},
\end{cases}
\quad t_s < t < t_e,
\label{eq:edge_gradient}
\end{equation}
yielding a linear ramp from~0 to~1 (or 1 to 0).
\end{definition}

The weights form a tensor
$W\!\in\!\mathbb{R}^{N\times N\times T}$ that captures change
intensity according to Eq.~\eqref{eq:edge_gradient}. The corresponding
procedure is summarized in Algorithm~\ref{alg:edgegrad}.

The interpolation procedure is $O(|E|T)$ and parallelisable across edges.

\begin{algorithm}[!h]
\caption{Edge-Gradient Interpolation}
\label{alg:edgegrad}
\KwIn{snapshots $\{E_t\}_{1:T}$}
\KwOut{tensor $W$ with intensities}

initialise $W_{uvt}\gets \mathbf{1}[e_{uv}^{(t)}=1]$\;

\ForEach{edge $(u,v)$}{
  identify all birth--death intervals $(t_s,t_e)$\;
  \For{$t\gets t_s+1$ \KwTo $t_e-1$}{
    \uIf{$e_{uv}^{(t_s)}=0$}{
      $W_{uvt}\gets (t-t_s)/(t_e-t_s)$
    }\Else{
      $W_{uvt}\gets 1-(t-t_s)/(t_e-t_s)$
    }
  }
}
\end{algorithm}

\subsection{EG\mbox{-}RL Architecture}
\label{subsec:architecture}

EG\mbox{-}RL follows the \emph{encoder--mixer--classifier} paradigm of
GraphMixer \cite{chen2023graphmixer} while extending it along two
dimensions: (i) explicit \textbf{edge-gradient signals}
\ref{subsec:edgegradient}; (ii) a \textbf{reinforcement loop}
\ref{subsec:rl} that adaptively selects the most informative
edge-gradient snapshots.  Fig.~\ref{fig:2} gives the full
pipeline; we detail each component below.

\paragraph{Fixed time encoding.}
Instead of learning a positional encoder, we adopt the \emph{cosine lattice}
used in GraphMixer. For any timestamp $t\in\mathbb{R}_{\ge 0}$ and encoding
dimension $d$, we define the fixed time encoding as
\begin{equation}
  \phi(t)=\bigl[\cos(t\,\omega_i)\bigr]_{i=1}^{d},
  \qquad
  \omega_i=\alpha^{-\frac{i-1}{\beta}},
  \label{eq:timeenc}
\end{equation}
where $\phi(t)\in\mathbb{R}^d$ and $\omega_i$ denotes the $i$-th frequency.
Following GraphMixer, we use the fixed setting $\alpha=\beta=\sqrt{d}$ and set
$d=100$ throughout. Here, $\alpha$ and $\beta$ control how the frequencies decay
across dimensions, while $d$ determines the size of the time encoding. This
parameterization yields a smoothly decaying bank of multi-scale frequencies:
lower-index channels vary faster and capture fine-grained temporal differences,
whereas higher-index channels vary more slowly and capture coarse temporal
patterns. Setting $d=100$ keeps the encoding sufficiently expressive while
maintaining modest computational overhead. This fixed encoding has two desirable
properties: (i) nearby timestamps yield highly correlated $\phi(t)$, supporting
interpolation; (ii) as $i$ grows, $\omega_i\to 0$, so higher-index channels vary
more smoothly and remain distinguishable even for large $t$.

\paragraph{Node \& link encoders.}
For each node $v$, we construct three components: a raw node feature
$x_v\in\mathbb{R}^{d_x}$, a one-hot identity embedding
$\mathbf{1}_v\in\{0,1\}^{|V|}$, and the fixed time encoding
$\phi(t)\in\mathbb{R}^{d}$. When no raw node attributes are available,
we use the zero vector for $x_v$. Since all three components are vectors,
they can be concatenated into a single input in
$\mathbb{R}^{d_x+|V|+d}$ and passed to a two-layer MLP:
\begin{equation}
  h_v^{(t)}
  = \mathrm{MLP}_n\!\bigl([x_v \,\|\, \mathbf{1}_v \,\|\, \phi(t)]\bigr),
  \label{eq:nodeenc}
\end{equation}
where the identity embedding provides a lightweight way to distinguish
nodes even when raw attributes are weak or unavailable.

For every edge $(u,v)$, we analogously combine its interpolated gradient
weight $w_{uv}^{(t)}\in\mathbb{R}$, optional raw edge feature
$x_{uv}\in\mathbb{R}^{d_e}$, and the same time encoding
$\phi(t)\in\mathbb{R}^{d}$. If edge attributes are unavailable, we use
the zero vector for $x_{uv}$. Their concatenation therefore lies in
$\mathbb{R}^{1+d_e+d}$ and is mapped by a one-layer MLP:
\begin{equation}
  \tilde{h}_{uv}^{(t)}
  = \mathrm{MLP}_\ell\!\bigl([w_{uv}^{(t)} \,\|\, x_{uv} \,\|\, \phi(t)]\bigr).
  \label{eq:edgeenc_local}
\end{equation}
Because nodes participate in variable-length interaction sequences, we
zero-pad to a fixed window $K$ (recent first) and average over
$k$ spatial hops:
\begin{equation}
  h_{uv}^{(t)}
  = \mathrm{MeanPool}_{\le k\text{ hops}}
  \bigl(\underbrace{[\tilde{h}_{uv}^{(t)},\dots]}_{\text{length}=K}\bigr),
  \label{eq:linkenc}
\end{equation}
where padding length also encodes activity frequency: sparsely
interacting nodes receive more zeros, a signal absent from
attention-based models.

\paragraph{Token/channel mixer.}
Stacking $L$ GraphMixer blocks, we alternately mix along the \emph{token}
(axis over nodes/edges) and \emph{channel} (feature) dimensions.  For
block $\ell$ with input matrix $Z\!\in\!\mathbb{R}^{N\times C}$:
\begin{align}
  Z'         &= Z
               + \mathrm{MLP}_{\!\text{token}}\!\bigl(\mathrm{LN}(Z)\bigr),
  \label{eq:tokenmix}\\
  Z_{\ell+1} &= Z'
               + \mathrm{MLP}_{\!\text{chan}}\!\bigl(\mathrm{LN}(Z')\bigr),
  \label{eq:chanmix}
\end{align}
where LN is pre-layer normalization and both MLPs use depth-wise linear
projections and $\mathrm{GELU}$ activations.  The residual structure
(\ref{eq:tokenmix})--(\ref{eq:chanmix}) preserves gradient flow and
requires \emph{no} self-attention, yielding
\emph{O}$(N\!+\!C)$ parameter cost per block.

\paragraph{Graph embedding and classifier.}
After $L$ blocks, we mean-pool the token matrix to obtain a compact graph-level
summary $g=\mathrm{Mean}(Z_L)$. For a candidate link $(u,v)$ at time $t$, we use
the endpoint node representations $h_u^{(t)}$ and $h_v^{(t)}$ from
Eq.~\eqref{eq:nodeenc}, together with the pair-specific edge representation
$h_{uv}^{(t)}$ from Eq.~\eqref{eq:linkenc}. A two-layer MLP with hidden size
$\tfrac12\dim(g)$ then outputs the link-existence logit:
\begin{equation}
\hat{y}_{uv}
=
\mathrm{MLP}_c\!\bigl([g \,\|\, h_u^{(t)} \,\|\, h_v^{(t)} \,\|\, h_{uv}^{(t)}]\bigr).
\label{eq:classifier}
\end{equation}

\paragraph{Interface to RL selector.}
During each RL epoch the selector proposes a set of candidate
edge-gradient snapshots.  Accepted edges simply enter
(\ref{eq:linkenc}); rejected ones are masked out by setting
$w_{uv}^{(t)}\!=\!0$ before the mixer.  Because time encoding
(\ref{eq:timeenc}) and padding are differentiable, gradients propagate
\emph{through} the selector's decisions, enabling policy updates via the critic reward.

\paragraph{Complexity.}
Let $|E|$ be the number of edges in the selected snapshot, $C$ the
feature width, and $L$ the mixer depth.  Node/edge encoders cost
$\mathcal{O}(|E|)$, each mixer block
$\mathcal{O}(|E|C)$, and RL sampling $\mathcal{O}(kT)$.  The overall
$\mathcal{O}(|E|LC + kT)$ is linear in both edges and snapshots.

\subsection{Reinforcement-Guided Edge-Gradient Selection}
\label{subsec:rl}

\paragraph{Markov decision process (MDP).}
Snapshot selection is framed as a Markov decision process,
$\mathcal{M}=(\mathcal{S},\mathcal{A},\mathcal{T},\mathcal{R})$.
At step $t$ the agent evaluates one candidate gradient
$e_t\!\in\!E_C$; state
$s_t=(G_t^{\text{aug}},e_t)$, action $a_t\!\in\!\{0,1\}$ and
$G_{t+1}^{\text{aug}}=G_t^{\text{aug}}\cup\{e_t\}$ if $a_t=1$.

\paragraph{Policy $\pi_\theta$ \& Critic $Q_\psi$.}
Both share graph embedding of $G_t^{\text{aug}}$ concatenated with an
edge embedding of $e_t$; each is a two-layer MLP.

\paragraph{Reward.}
After an episode we fine-tune the classifier for one epoch with
\emph{encoder frozen} and compute validation AUC.
Reward is the relative gain
$r=
  \bigl(\mathrm{AUC}(G_T^{\text{aug}})-\mathrm{AUC}(G)\bigr)
/\mathrm{AUC}(G)$, clipped to $[-1,1]$.

\paragraph{Updates.}
REINFORCE with baseline:
\begin{equation}
\begin{aligned}
\nabla_\theta J
= \sum_t \bigl(R_t - Q_\psi(s_t)\bigr)\nabla_\theta \log \pi_\theta(a_t \mid s_t).
\end{aligned}
\end{equation}
$Q_\psi$ minimises TD(0); both use Adam. Section~\ref{subsec:terminal_extension}
introduces a distinct terminal-reward extension; it does not replace this
relative-reward, TD(0) formulation.
\subsection{Training Objective and Complexity}
\label{subsec:training}
The joint loss
\[
  \mathcal{L}
  =\mathcal{L}_{\text{BCE}}
  +\lambda \mathcal{L}_{\text{RL}}
  +\beta\|\Theta\|_2^{2}
\]
combines supervised BCE, policy gradient loss and
$L_2$ regularisation.  
Forward/ backward cost is $O(|E|L)$ for the mixer plus
$O(kT)$ for RL sampling---\emph{linear} in edges and snapshots---making
EG-RL more scalable than message-passing temporal GNNs.
\subsection{Terminal-Reward Training Extension}
\label{subsec:terminal_extension}
Algorithm~\ref{alg:train} gives the extended training loop. It uses the
signed, unnormalized validation AUC difference (not its absolute value)
and regresses the critic to the episode's terminal reward, without TD
bootstrapping. This differs from the relative-AUC and TD(0) formulation
in Section~\ref{subsec:rl}; the two configurations must be identified
separately in comparisons. The reward-metric study varies AUC/AP signals,
not these critic objectives.

\begin{algorithm}[!h]
\caption{Terminal-Reward EG-RL Training Extension}
\label{alg:train}
\KwIn{dynamic graph $\mathcal{G}_{1:T}$, epochs $N_{\text{ep}}$, policy $\pi_\theta$, critic $Q_\psi$}
\KwOut{trained parameters $\Theta,\theta,\psi$}

\For{$\text{epoch}=1$ \KwTo $N_{\text{ep}}$}{
  \vspace{2pt}
  \tcp*{\small Stage 0: edge-gradient interpolation}
  $W \leftarrow$ \textsc{EdgeGradientAugmentation}$(\mathcal{G}_{1:T})$\;

  \vspace{2pt}
  \tcp*{\small Stage I: RL sampling of augmented snapshot}
  $G^{\text{aug}}\leftarrow G;\;\; \mathcal{B}\leftarrow\varnothing$\;
  \ForEach{candidate edge $e_t\in E_C$}{
      $s_t\leftarrow\bigl(G^{\text{aug}},\,e_t\bigr)$\;
      sample $a_t\!\sim\!\pi_\theta(\cdot\mid s_t)$\;
      \If(\tcp*[f]{keep edge}){$a_t=1$}{
        insert $e_t$ with weight $w_{e_t}$ into $G^{\text{aug}}$\;
      }
      store $(s_t,a_t)$ in buffer $\mathcal{B}$\;
  }

  \vspace{2pt}
  \tcp*{\small Stage II: supervised update (encoder $\Rightarrow$ mixer)}
  compute logits $\hat{y}$ on $G^{\text{aug}}$ via Eq.\,\eqref{eq:nodeenc}--\eqref{eq:chanmix}\;
  $\mathcal{L}_{\text{BCE}}\leftarrow\mathrm{BCE}\bigl(\hat{y},y\bigr)$\;
  update encoder \& mixer params $\Theta$ with Adam on $\mathcal{L}_{\text{BCE}}$\;

  \vspace{2pt}
  \tcp*{\small Stage III: one-step fine-tune \& reward}
  freeze $\Theta$; fine-tune classifier 1 epoch on $G^{\text{aug}}$\;
  $r\leftarrow\mathrm{AUC}\bigl(G^{\text{aug}}\bigr)-\mathrm{AUC}(G)$\;

  \vspace{2pt}
  \tcp*{\small Stage IV: policy \& critic update (REINFORCE)}
  \ForEach{$(s_t,a_t)\in\mathcal{B}$}{
      $\delta\leftarrow r-Q_\psi(s_t)$\;
      $\theta\leftarrow\theta+\eta_\pi\,\delta\,\nabla_\theta\!\log\pi_\theta(a_t\mid s_t)$\;
      $\psi\leftarrow\psi-\eta_Q\,\nabla_\psi\!\bigl(Q_\psi(s_t)-r\bigr)^{\!2}$\;
  }
}
\end{algorithm}

\section{Experiments}\label{sec:exp}

We benchmark EG--RL on \textbf{8} public
dynamic-graph datasets and compare with \textbf{12} established
baselines spanning recurrent, attention and mixer families. The main prediction results are averaged over three seeds; numbers are reported as \emph{mean\,$\pm$\,s.d}.

\subsection{Datasets}
\label{subsec:datasets}
We evaluate on eight widely-used temporal graph benchmark datasets curated by
Yu \emph{et al.}~\cite{yu2023towards}, covering diverse domains including social,
interaction, transportation, and proximity networks. Specifically,
\textbf{Wikipedia}, \textbf{Reddit}, and \textbf{MOOC} are bipartite
interaction graphs with edge features; \textbf{LastFM} models user--song
listening events; \textbf{Enron} captures email communications;
\textbf{Social Evolution} records mobile proximity with low-dimensional link
features; \textbf{UCI} is an online communication network; and
\textbf{Flights} tracks airport traffic during COVID-19.

\begin{table*}[t]
\centering
\small
\caption{Statistics of the datasets used in experiments. \#N\&L Feat denotes the dimensions of node and link features.}
\label{tab:dataset_stats}
\setlength{\tabcolsep}{6pt}
\begin{tabular*}{\textwidth}{@{\extracolsep{\fill}} lccccccc}
\toprule
Dataset & Domain & \#Nodes & \#Links & \#N\&L Feat & Bipartite & Duration & Time Granularity \\
\midrule
Wikipedia    & Social      & 9,227  & 157,474   & -- \& 172 & True  & 1 month   & Unix timestamps \\
Reddit       & Social      & 10,984 & 672,447   & -- \& 172 & True  & 1 month   & Unix timestamps \\
MOOC         & Interaction & 7,144  & 411,749   & -- \& 4   & True  & 17 months & Unix timestamps \\
LastFM       & Interaction & 1,980  & 1,293,103 & -- \& --  & True  & 1 month   & Unix timestamps \\
Enron        & Social      & 184    & 125,235   & -- \& --  & False & 3 years   & Unix timestamps \\
Social Evo.  & Proximity   & 74     & 2,099,519 & -- \& 2   & False & 8 months  & Unix timestamps \\
UCI          & Social      & 1,899  & 59,835    & -- \& --  & False & 196 days  & Unix timestamps \\
Flights      & Transport   & 13,169 & 1,927,145 & -- \& 1   & False & 4 months  & days \\
\bottomrule
\end{tabular*}
\end{table*}

\subsection{Baselines}
We compare twelve baselines: JODIE \cite{kumar2019predicting}, DyRep
\cite{dyrep}, and TGN \cite{tgn} for recurrent/event-state modelling;
TGAT \cite{xuinductive}, CAWN \cite{cawn}, TCL \cite{wang2021tcl}, and
DyGFormer \cite{yu2023towards} for temporal attention or walk encoding;
NAT \cite{nat}, PINT \cite{pint}, GraphMixer \cite{chen2023graphmixer}, and
TPNet \cite{tpnet} for neighborhood, expressive, mixer, or walk-matrix
representations; and the EdgeBank heuristic \cite{poursafaei2022towards}.
Recent methods discussed in Section II are not additional evaluated baselines.

TPNet~\cite{tpnet} uses time-decayed temporal-walk matrices and
random-feature propagation. EG-RL instead constructs edge-transition
weights and learns which causally available augmentation candidates to retain.

\providecommand{\pms}[1]{\,{\scriptsize$\pm$#1}}

These baselines cover a broad spectrum of approaches and represent strong competitors for temporal link prediction.
\subsection{Experimental Setup and Reproducibility}
\label{subsec:implementation}
The main benchmark experiments are implemented in PyTorch 2.0 with PyTorch Geometric 2.3 and \texttt{torch\_scatter}. Encoder and mixer parameters are optimized with Adam (learning rate $1\times10^{-3}$, dropout 0.1, weight decay $5\times10^{-5}$), while the policy and critic use separate Adam optimizers with a learning rate of $3\times10^{-4}$. Temporal edges are time-sorted and processed in mini-batches of 2{,}000; each positive event is paired with one on-the-fly negative during training and ranked against 100 degree-matched negatives at test time. The RL selector examines at most $0.1\,|E_t|$ candidate gradients per snapshot (capped at 512). We report results averaged over three random seeds and release anonymous code and processed datasets at \url{https://anonymous.4open.science/r/EG-RL-671D}.

\paragraph{Compute resources.}
The reported training workstation has an Intel i7-13700K CPU, 32\,GB RAM,
and one RTX 5090 GPU (32\,GB). A Reddit epoch takes approximately
12 seconds with 5.3\,GB peak GPU memory. Preliminary tuning and ablations
require additional compute.

\begin{figure}[t]
  \centering
  \begin{subfigure}[b]{0.49\columnwidth}
    \centering
    \begin{minipage}[c][1.62in][c]{\linewidth}
      \centering
      \includegraphics[width=\linewidth,height=1.62in,keepaspectratio,trim={9.6bp 9.6bp 6bp 8.4bp},clip]{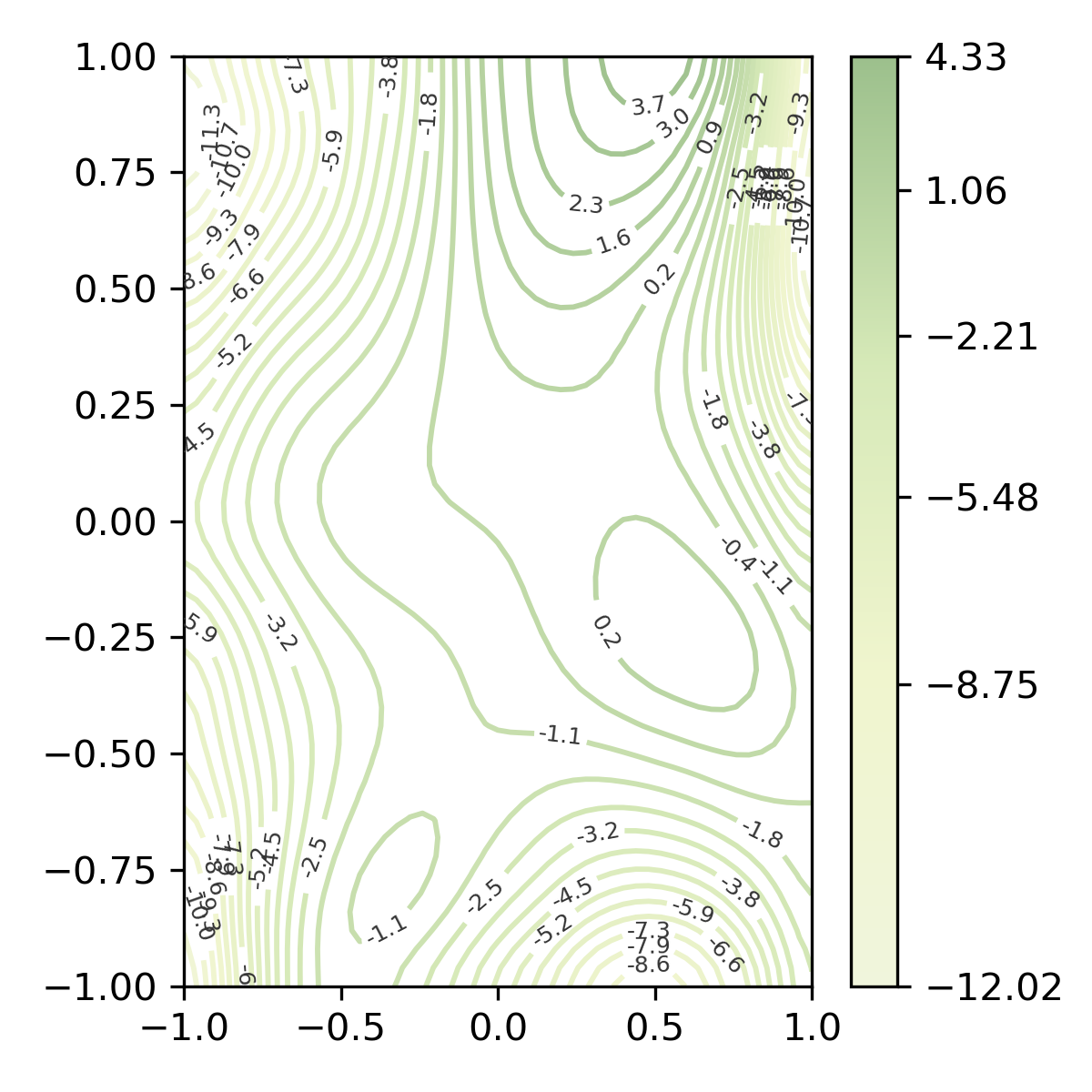}
    \end{minipage}
    \caption{DyGFormer (2D)}
  \end{subfigure}\hfill
  \begin{subfigure}[b]{0.49\columnwidth}
    \centering
    \begin{minipage}[c][1.62in][c]{\linewidth}
      \centering
      \includegraphics[width=\linewidth,height=1.62in,keepaspectratio,trim={60bp 4.8bp 25.2bp 30bp},clip]{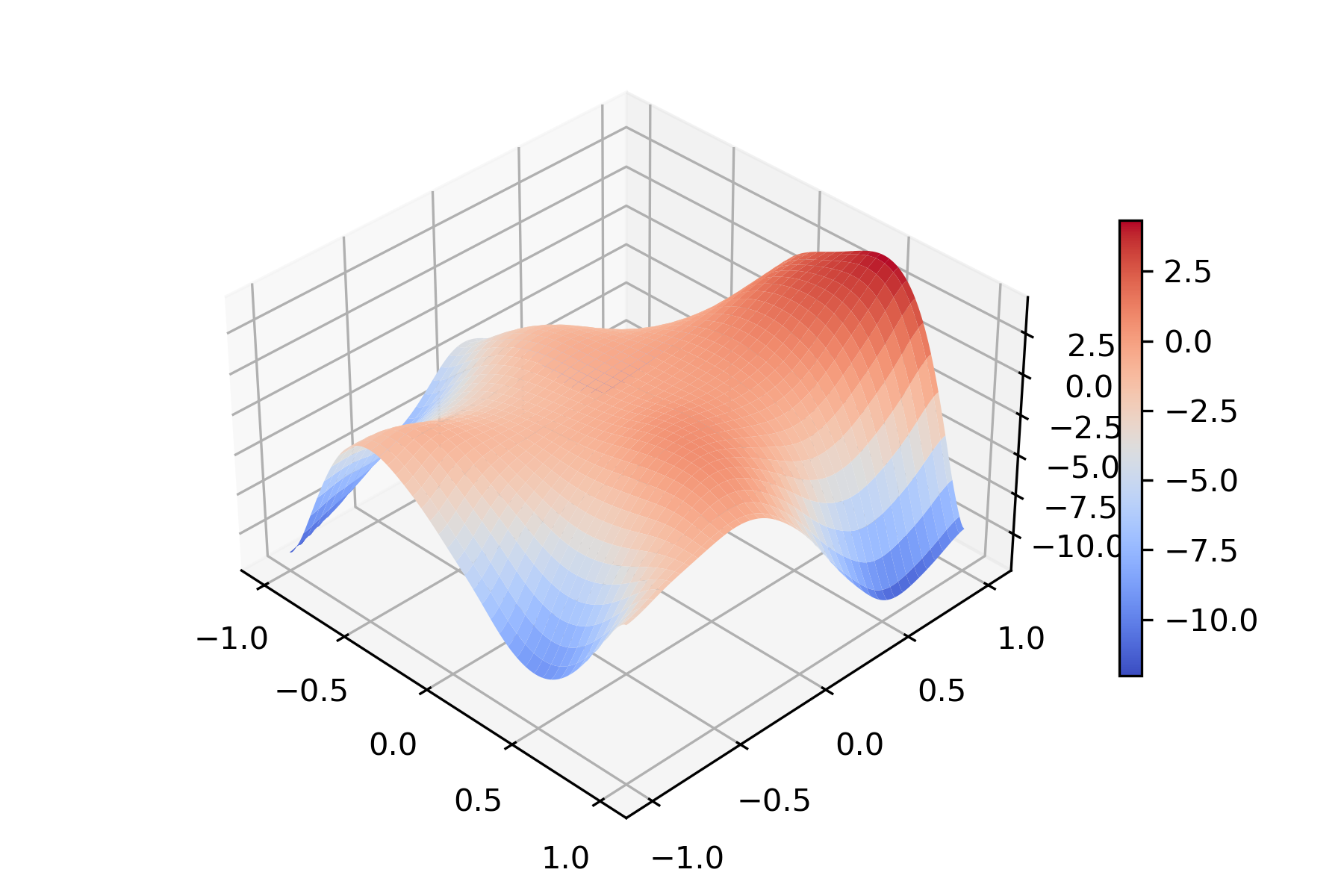}
    \end{minipage}
    \caption{DyGFormer (3D)}
  \end{subfigure}
  \par\vspace{0.7ex}
  \begin{subfigure}[b]{0.49\columnwidth}
    \centering
    \begin{minipage}[c][1.62in][c]{\linewidth}
      \centering
      \includegraphics[width=\linewidth,height=1.62in,keepaspectratio,trim={9.6bp 9.6bp 6bp 8.4bp},clip]{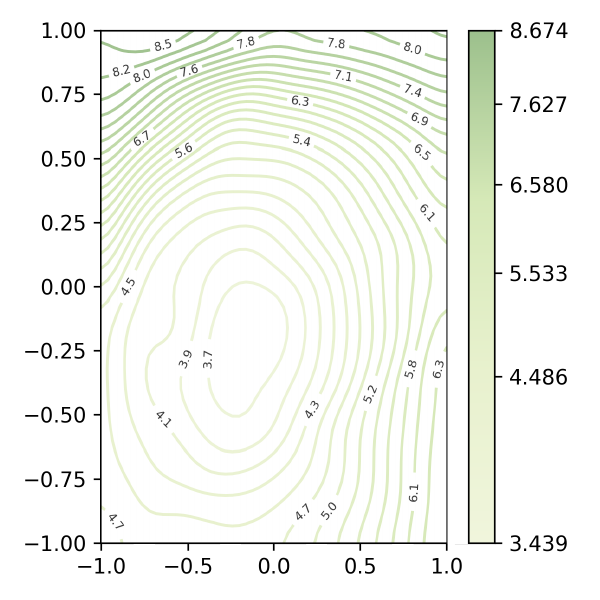}
    \end{minipage}
    \caption{EG-RL (2D)}
  \end{subfigure}\hfill
  \begin{subfigure}[b]{0.49\columnwidth}
    \centering
    \begin{minipage}[c][1.62in][c]{\linewidth}
      \centering
      \includegraphics[width=\linewidth,height=1.62in,keepaspectratio,trim={60bp 4.8bp 25.2bp 30bp},clip]{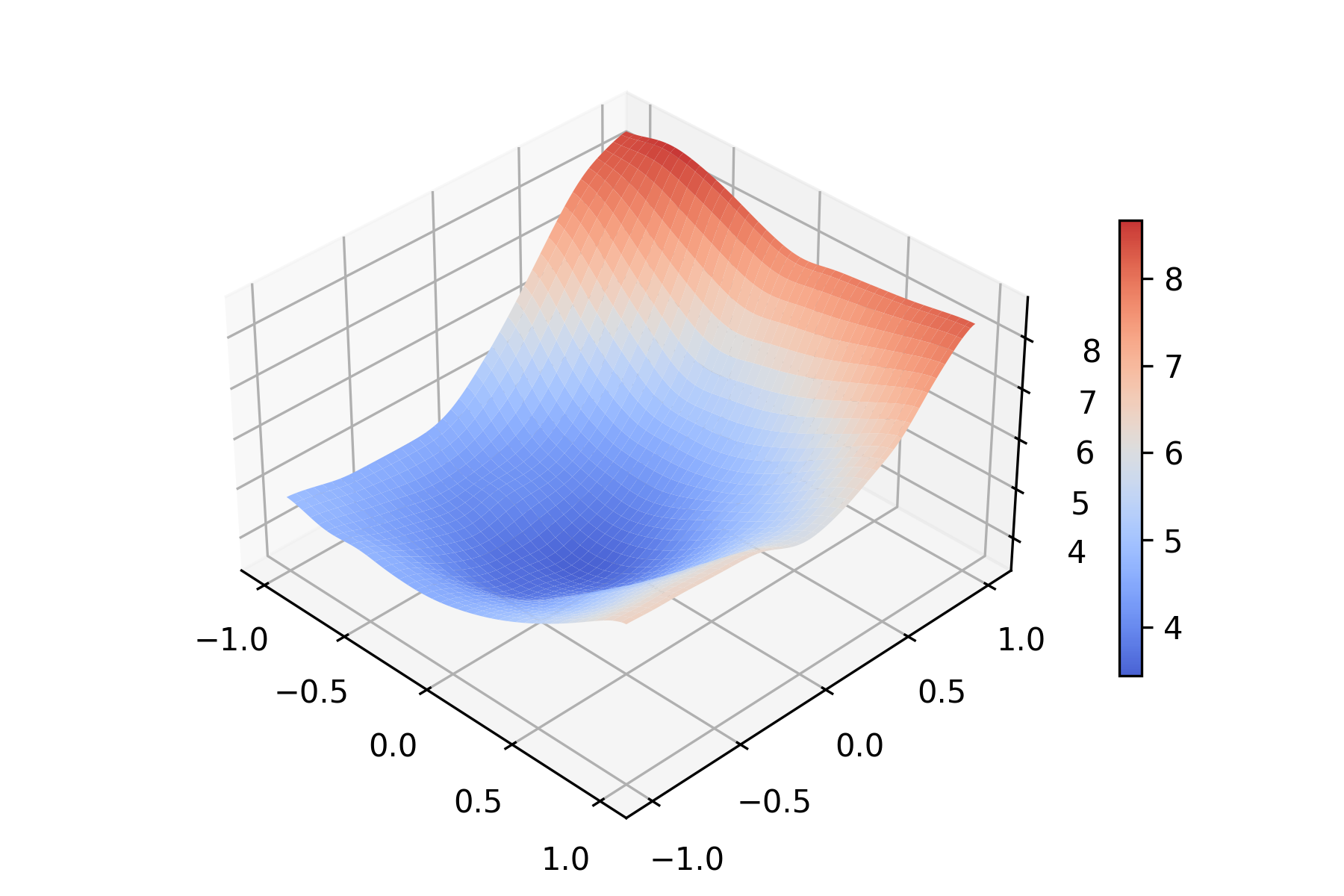}
    \end{minipage}
    \caption{EG-RL (3D)}
  \end{subfigure}
  \par\vspace{0.7ex}
  \begin{subfigure}[b]{0.49\columnwidth}
    \centering
    \begin{minipage}[c][1.62in][c]{\linewidth}
      \centering
      \includegraphics[width=\linewidth,height=1.62in,keepaspectratio,trim={9.6bp 9.6bp 6bp 8.4bp},clip]{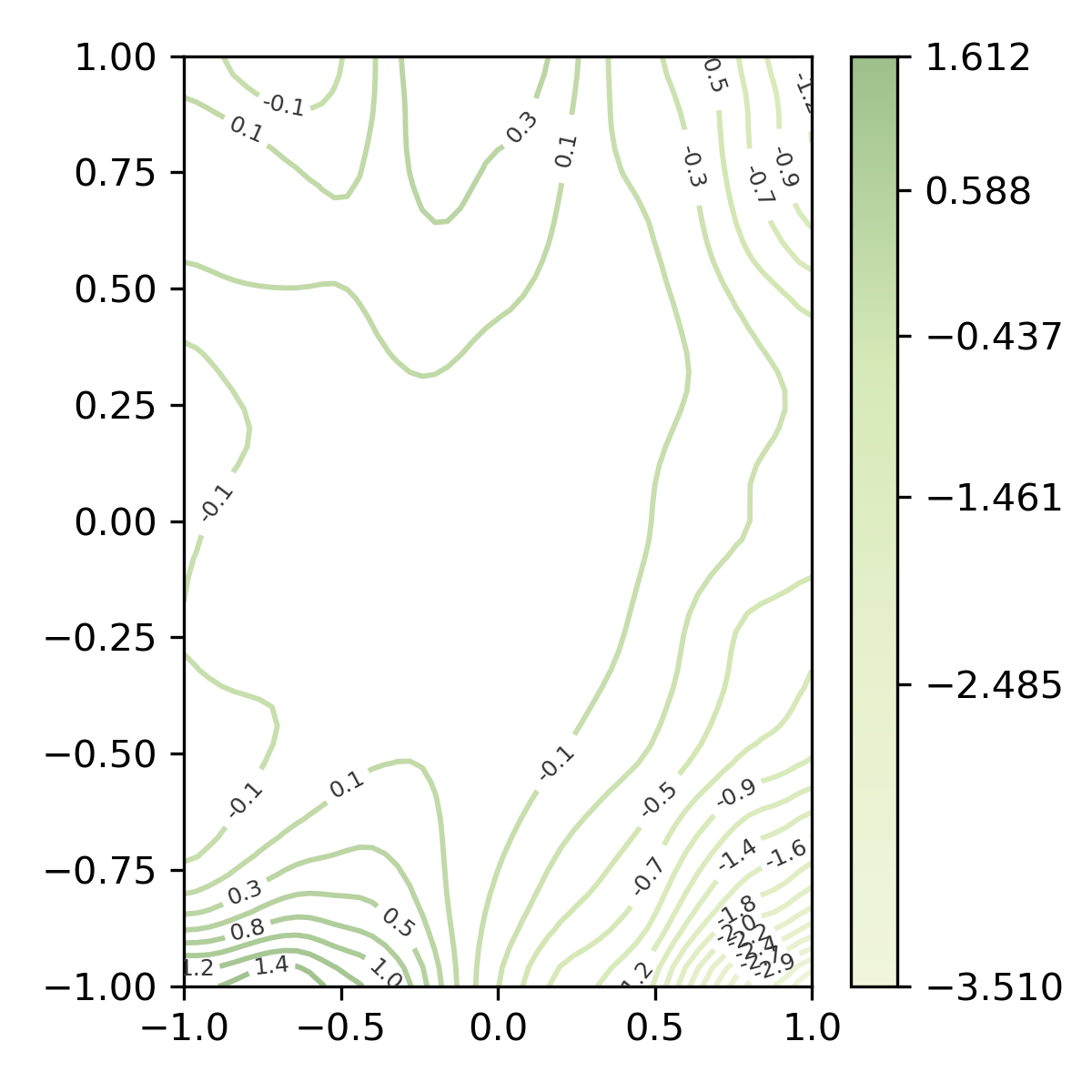}
    \end{minipage}
    \caption{TPNet (2D)}
  \end{subfigure}\hfill
  \begin{subfigure}[b]{0.49\columnwidth}
    \centering
    \begin{minipage}[c][1.62in][c]{\linewidth}
      \centering
      \includegraphics[width=\linewidth,height=1.62in,keepaspectratio,trim={60bp 4.8bp 25.2bp 30bp},clip]{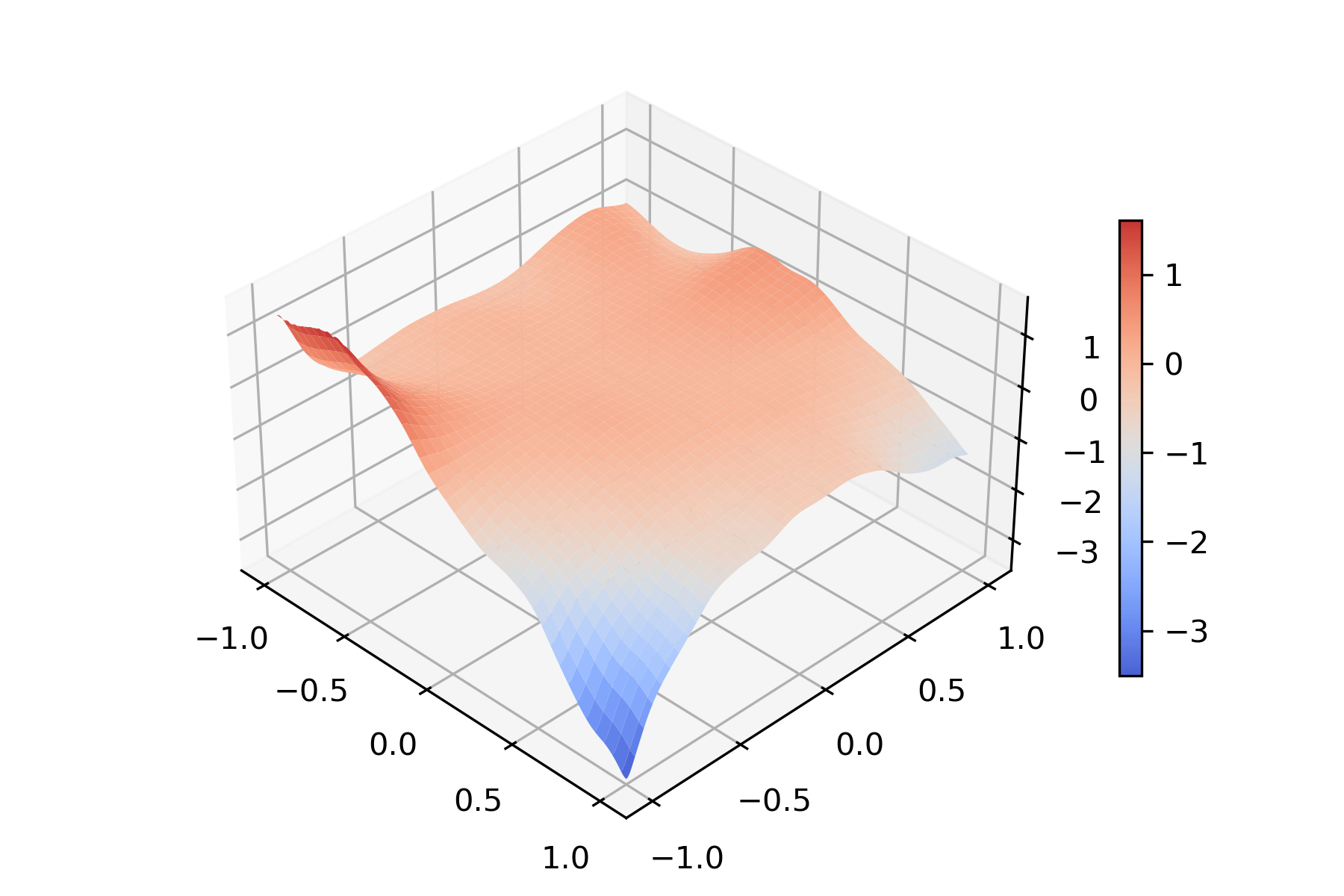}
    \end{minipage}
    \caption{TPNet (3D)}
  \end{subfigure}
  \caption{Comparison of loss landscapes on Wikipedia. Rows show DyGFormer, EG-RL, and TPNet, respectively. Left: 2D contour plots; right: 3D surfaces.}
  \label{fig:loss_landscape_wiki}
\end{figure}

\subsection{Main Empirical Results}

Table~\ref{tab:main} reports the AP and AUC results of EG-RL and twelve strong baselines on eight temporal-graph benchmarks. Overall, EG-RL achieves the best average rank in both metrics and outperforms competing methods on most datasets.

In particular, EG-RL shows especially strong performance on Wikipedia and Reddit, where it achieves the best AP and AUC among all compared methods. It also delivers clear improvements on sparse datasets such as MOOC and UCI, suggesting that edge-gradient augmentation is particularly beneficial when temporal interactions are limited. Even on denser datasets such as Flights, EG-RL remains highly competitive and consistently ranks among the top-performing methods.

The final row of Table~\ref{tab:main} further confirms this trend: EG-RL attains an average rank of 1.75 in AP and 1.88 in AUC, outperforming all baselines overall. These results demonstrate the effectiveness of combining edge-gradient modeling with reinforcement-guided snapshot selection for temporal link prediction.
\begin{table*}[t]
\centering
\small
\setlength{\tabcolsep}{3pt}
\caption{Temporal link prediction on 8 benchmarks. Best result in \textbf{bold}; runner-up \underline{underlined}.}
\label{tab:main}
\resizebox{\textwidth}{!}{%
\begin{tabular}{@{}llccccccccccccc@{}}
\toprule
\multicolumn{2}{c}{Metric} & JODIE & DyRep & TGAT & TGN & CAWN & EdgeBank & TCL & GraphMixer & NAT & PINT & DyGFormer & TPNet & EG-RL \\
\midrule
\multirow{8}{*}{\rotatebox{90}{\textbf{AP} (\%)}} 
 & Wikipedia & 96.50\pms{0.14} & 94.86\pms{0.06} & 96.94\pms{0.06} & 98.45\pms{0.06} & 98.76\pms{0.03} & 90.37\pms{0.00} & 96.47\pms{0.16} & 97.25\pms{0.03} & 98.03\pms{0.07} & 98.45\pms{0.04} & 99.03\pms{0.02} & \underline{99.32}\pms{0.03} & \textbf{99.99}\pms{0.00} \\
 & Reddit    & 98.31\pms{0.14} & 98.22\pms{0.04} & 98.52\pms{0.02} & 98.63\pms{0.06} & 99.11\pms{0.01} & 94.86\pms{0.00} & 97.53\pms{0.02} & 97.31\pms{0.01} & 99.13\pms{0.10} & 99.15\pms{0.02} & 99.22\pms{0.01} & \underline{99.27}\pms{0.00} & \textbf{99.99}\pms{0.00} \\
 & MOOC      & 80.23\pms{2.44} & 81.97\pms{0.49} & 85.84\pms{0.15} & 89.15\pms{1.60} & 80.15\pms{0.25} & 57.97\pms{0.00} & 82.38\pms{0.24} & 82.78\pms{0.15} & 85.88\pms{0.55} & 88.08\pms{0.86} & 87.52\pms{0.49} & \underline{96.39}\pms{0.09} & \textbf{99.99}\pms{0.00} \\
 & LastFM    & 70.85\pms{2.13} & 71.92\pms{2.21} & 73.42\pms{0.21} & 77.07\pms{3.97} & 86.99\pms{0.06} & 79.29\pms{0.00} & 67.27\pms{2.16} & 75.61\pms{0.24} & 88.02\pms{1.94} & 89.66\pms{1.81} & 93.00\pms{0.12} & \underline{94.50}\pms{0.08} & \textbf{96.69}\pms{0.01} \\
 & Enron     & 84.77\pms{0.30} & 82.38\pms{3.36} & 71.12\pms{0.97} & 86.53\pms{1.11} & 89.56\pms{0.09} & 83.53\pms{0.00} & 79.70\pms{0.71} & 82.25\pms{0.16} & 90.60\pms{0.66} & 92.20\pms{0.15} & \underline{92.47}\pms{0.12} & \textbf{92.90}\pms{0.17} & 92.41\pms{0.05} \\
 & Social Evo. & 89.89\pms{0.55} & 88.87\pms{0.30} & 93.16\pms{0.17} & 93.57\pms{0.17} & 84.96\pms{0.09} & 74.95\pms{0.00} & 93.13\pms{0.16} & 93.37\pms{0.07} & 88.92\pms{3.45} & 94.42\pms{0.03} & \textbf{94.73}\pms{0.01} & \textbf{94.73}\pms{0.02} & 94.54\pms{0.04} \\
 & UCI       & 89.43\pms{1.09} & 65.14\pms{2.30} & 79.63\pms{0.70} & 92.34\pms{1.04} & 95.18\pms{0.06} & 76.20\pms{0.00} & 89.57\pms{1.63} & 93.25\pms{0.57} & 93.40\pms{0.26} & 96.45\pms{0.11} & 95.79\pms{0.17} & \textbf{97.35}\pms{0.04} & \underline{96.63}\pms{0.04} \\
 & Flights   & 95.60\pms{1.73} & 95.29\pms{0.72} & 94.03\pms{0.18} & 97.95\pms{0.14} & 98.51\pms{0.01} & 89.35\pms{0.00} & 91.23\pms{0.02} & 90.99\pms{0.05} & 98.57\pms{0.12} & 98.80\pms{0.02} & \underline{98.91}\pms{0.01} & 97.73\pms{0.02} & \textbf{98.86}\pms{0.03} \\
 \addlinespace[2pt]
 \multicolumn{2}{r}{Avg. Rank} & 9.63 & 10.75 & 9.38 & 6.13 & 7.00 & 13.00 & 10.50 & 9.13 & 6.00 & 3.88 & 2.75 & \underline{2.25} & \textbf{1.75} \\
\midrule
\multirow{8}{*}{\rotatebox{90}{\textbf{AUC} (\%)}} 
 & Wikipedia & 96.33\pms{0.07} & 94.37\pms{0.09} & 96.67\pms{0.07} & 98.37\pms{0.07} & 98.54\pms{0.04} & 90.78\pms{0.00} & 95.84\pms{0.18} & 96.92\pms{0.03} & 97.75\pms{0.11} & 98.16\pms{0.06} & 98.91\pms{0.02} & \underline{99.30}\pms{0.02} & \textbf{99.99}\pms{0.00} \\
 & Reddit    & 98.31\pms{0.05} & 98.17\pms{0.05} & 98.47\pms{0.02} & 98.60\pms{0.06} & 99.01\pms{0.01} & 95.37\pms{0.00} & 97.42\pms{0.02} & 97.17\pms{0.02} & 99.09\pms{0.10} & 99.09\pms{0.03} & 99.15\pms{0.01} & \underline{99.22}\pms{0.00} & \textbf{99.99}\pms{0.00} \\
 & MOOC      & 83.81\pms{2.09} & 85.03\pms{0.58} & 87.11\pms{0.19} & 91.21\pms{1.15} & 80.38\pms{0.26} & 60.86\pms{0.00} & 83.12\pms{0.18} & 84.01\pms{0.17} & 87.42\pms{0.58} & 90.55\pms{0.43} & 87.91\pms{0.58} & \underline{97.17}\pms{0.08} & \textbf{99.99}\pms{0.00} \\
 & LastFM    & 70.49\pms{1.66} & 71.16\pms{1.89} & 71.59\pms{0.18} & 78.47\pms{2.94} & 85.92\pms{0.10} & 83.77\pms{0.00} & 64.06\pms{1.16} & 73.53\pms{0.12} & 86.92\pms{2.72} & 89.28\pms{1.63} & 93.05\pms{0.10} & \underline{94.39}\pms{0.04} & \textbf{97.71}\pms{0.02} \\
 & Enron     & 87.96\pms{0.52} & 84.89\pms{3.00} & 68.89\pms{1.10} & 88.32\pms{0.99} & 90.45\pms{0.14} & 87.05\pms{0.00} & 75.74\pms{0.72} & 84.38\pms{0.21} & 91.68\pms{0.83} & 92.87\pms{0.34} & \underline{93.33}\pms{0.13} & \textbf{93.98}\pms{0.26} & 93.05\pms{0.05} \\
 & Social Evo. & 92.05\pms{0.46} & 90.76\pms{0.21} & 94.76\pms{0.16} & 95.39\pms{0.17} & 87.34\pms{0.08} & 81.60\pms{0.00} & 94.84\pms{0.17} & 95.23\pms{0.07} & 90.84\pms{3.72} & 96.16\pms{0.02} & \underline{96.30}\pms{0.01} & \textbf{96.43}\pms{0.02} & 95.10\pms{0.01} \\
 & UCI       & 90.44\pms{0.49} & 68.77\pms{2.34} & 78.53\pms{0.74} & 92.03\pms{1.13} & 93.87\pms{0.08} & 77.30\pms{0.00} & 87.82\pms{1.36} & 91.81\pms{0.67} & 92.31\pms{0.37} & 95.57\pms{0.16} & 94.49\pms{0.26} & \underline{96.79}\pms{0.05} & \textbf{96.81}\pms{0.03} \\
 & Flights   & 96.21\pms{1.42} & 95.95\pms{0.62} & 94.13\pms{0.17} & 98.22\pms{0.13} & 98.45\pms{0.01} & 90.23\pms{0.00} & 91.21\pms{0.02} & 91.13\pms{0.01} & 98.69\pms{0.10} & 98.89\pms{0.02} & \underline{98.93}\pms{0.01} & 97.75\pms{0.02} & \textbf{98.95}\pms{0.05} \\
 \addlinespace[2pt]
 \multicolumn{2}{r}{Avg. Rank} & 9.38 & 10.50 & 9.50 & 5.88 & 7.00 & 13.00 & 10.75 & 9.25 & 5.88 & 3.88 & 3.00 & \underline{2.38} & \textbf{1.88} \\
\bottomrule
\end{tabular}
}
\end{table*}
\begin{table*}[t]
\centering
\footnotesize
\caption{Ablation results on eight representative datasets (AP/AUC, \%).}
\label{tab:ablation}
\setlength{\tabcolsep}{3pt}
\resizebox{\textwidth}{!}{%
\begin{tabular}{@{}lcccccccc@{}}
\toprule
\multirow{2}{*}{Variant} &
\multicolumn{1}{c}{Wikipedia} &
\multicolumn{1}{c}{Reddit} &
\multicolumn{1}{c}{MOOC} &
\multicolumn{1}{c}{LastFM} &
\multicolumn{1}{c}{Enron} &
\multicolumn{1}{c}{Social Evo.} &
\multicolumn{1}{c}{Flights} &
\multicolumn{1}{c}{UCI} \\
 & AP/AUC & AP/AUC & AP/AUC & AP/AUC & AP/AUC & AP/AUC & AP/AUC & AP/AUC \\
\midrule
Full EG-RL      & \textbf{99.99 / 99.99} & \textbf{99.99 / 99.99} & \textbf{99.99 / 99.99} & \textbf{96.69 / 97.71} & \textbf{92.41 / 93.05} & \textbf{94.54 / 95.10} & \textbf{96.63 / 96.81} & \textbf{98.86 / 98.95} \\
w/o RL Selector & 98.46 / 98.59 & 98.91 / 99.05 & 88.17 / 91.21 & 93.12 / 93.87 & 93.42 / 91.08 & 93.95 / 94.81 & 75.11 / 96.08 & 95.82 / 90.11 \\
w/o Gradient    & 97.25 / 96.92 & 97.31 / 97.17 & 82.78 / 84.01 & 75.61 / 73.53 & 82.25 / 84.38 & 93.37 / 95.23 & 93.25 / 91.81 & 90.99 / 91.13 \\
Fixed $k$       & 98.22 / 98.15 & 98.44 / 98.71 & 85.09 / 89.77 & 92.64 / 93.20 & 92.14 / 90.31 & 93.12 / 94.02 & 73.26 / 95.03 & 94.17 / 88.60 \\
\bottomrule
\end{tabular}%
}
\end{table*}
\subsection{Ablation Study}
\label{subsec:ablation}

To quantify the contribution of each design choice, we compare
EG-RL with three simplified variants:

\begin{enumerate}
  \item \textbf{w/o Gradient}\,: remove edge--gradient interpolation and
        use binary edges.
  \item \textbf{w/o RL Selector}\,: keep all interpolated edges, i.e.,
        disable the actor--critic module.
  \item \textbf{Fixed $k$}\,: keep RL but set the candidate window
        radius to a fixed $k=1$, removing adaptivity.
\end{enumerate}

Table~\ref{tab:ablation} reports results on eight representative datasets that differ in scale and domain. All models share identical hyperparameters; we repeat each run with three seeds and report mean\,$\pm$\,std.

\paragraph{Findings.}
\begin{itemize}
\item \textbf{Edge gradients are critical.} Removing gradient interpolation reduces AP on Wikipedia from 99.99\% to 97.25\% ($-2.74$ percentage points) and on Flights from 96.63\% to 93.25\% ($-3.38$ points), confirming that continuous edge-change signals substantially improve prediction.
  \item \textbf{Adaptive selection matters.} Disabling the RL selector drops AP on Wikipedia by 1.53 points (99.99\% $\rightarrow$ 98.46\%) and on Flights by 21.52 points (96.63\% $\rightarrow$ 75.11\%), indicating that many interpolated edges introduce noise without learned filtering.
  \item \textbf{Window adaptivity helps.} Fixing the candidate window to $k=1$ reduces AP on Wikipedia by 1.77 points and on Flights by 23.37 points, showing that a dynamic selection radius is necessary to handle varying temporal densities.
\end{itemize}
These findings verify that both components---explicit edge-gradient encoding and adaptive RL-based selection---are indispensable to EG-RL's state-of-the-art performance.

\subsection{Loss Landscape Analysis}
\label{subsec:loss}

\begin{figure*}[t]
  \centering
  \begin{minipage}[t]{\columnwidth}
    \vspace{0pt}\centering
    \captionsetup{type=figure}
    \setcounter{subfigure}{0}
    \begin{subfigure}[b]{0.49\linewidth}
      \centering
      \begin{minipage}[c][1.45in][c]{\linewidth}
        \centering
        \includegraphics[width=\linewidth,height=1.45in,keepaspectratio,trim={6bp 6bp 2.4bp 4.8bp},clip]{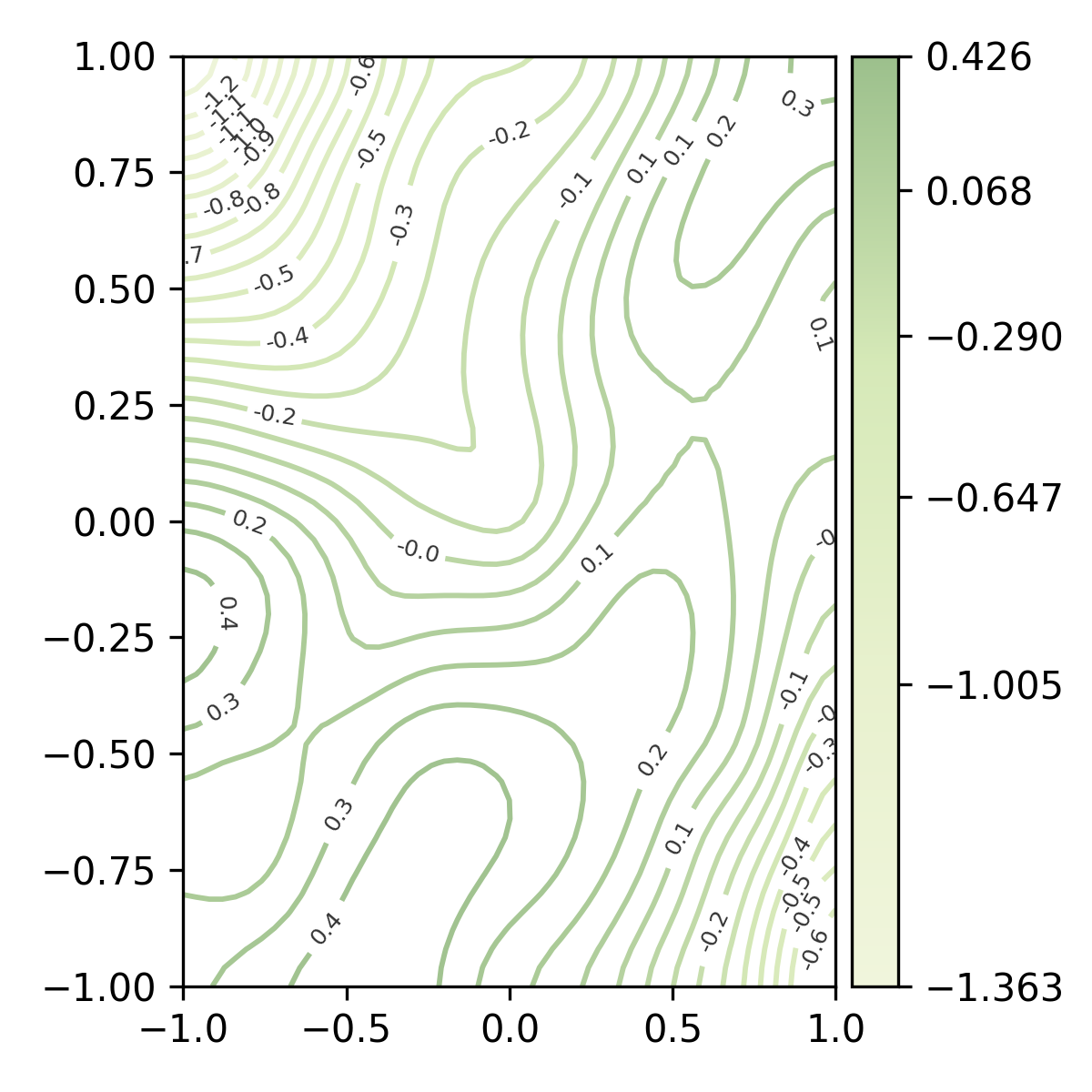}
      \end{minipage}
      \caption{ENRON}
    \end{subfigure}\hfill
    \begin{subfigure}[b]{0.49\linewidth}
      \centering
      \begin{minipage}[c][1.45in][c]{\linewidth}
        \centering
        \includegraphics[width=\linewidth,height=1.45in,keepaspectratio,trim={6bp 6bp 2.4bp 4.8bp},clip]{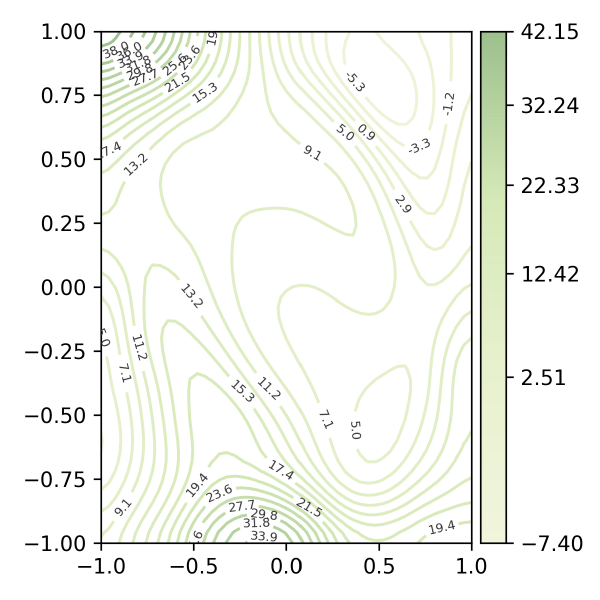}
      \end{minipage}
      \caption{LASTFM}
    \end{subfigure}
    \par\vspace{0.5ex}
    \begin{subfigure}[b]{0.49\linewidth}
      \centering
      \begin{minipage}[c][1.45in][c]{\linewidth}
        \centering
        \includegraphics[width=\linewidth,height=1.45in,keepaspectratio,trim={6bp 6bp 2.4bp 4.8bp},clip]{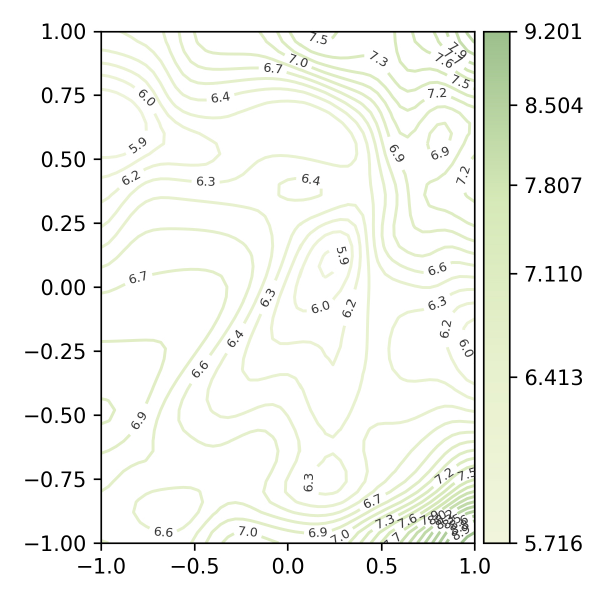}
      \end{minipage}
      \caption{MOOC}
    \end{subfigure}\hfill
    \begin{subfigure}[b]{0.49\linewidth}
      \centering
      \begin{minipage}[c][1.45in][c]{\linewidth}
        \centering
        \includegraphics[width=\linewidth,height=1.45in,keepaspectratio,trim={6bp 6bp 2.4bp 4.8bp},clip]{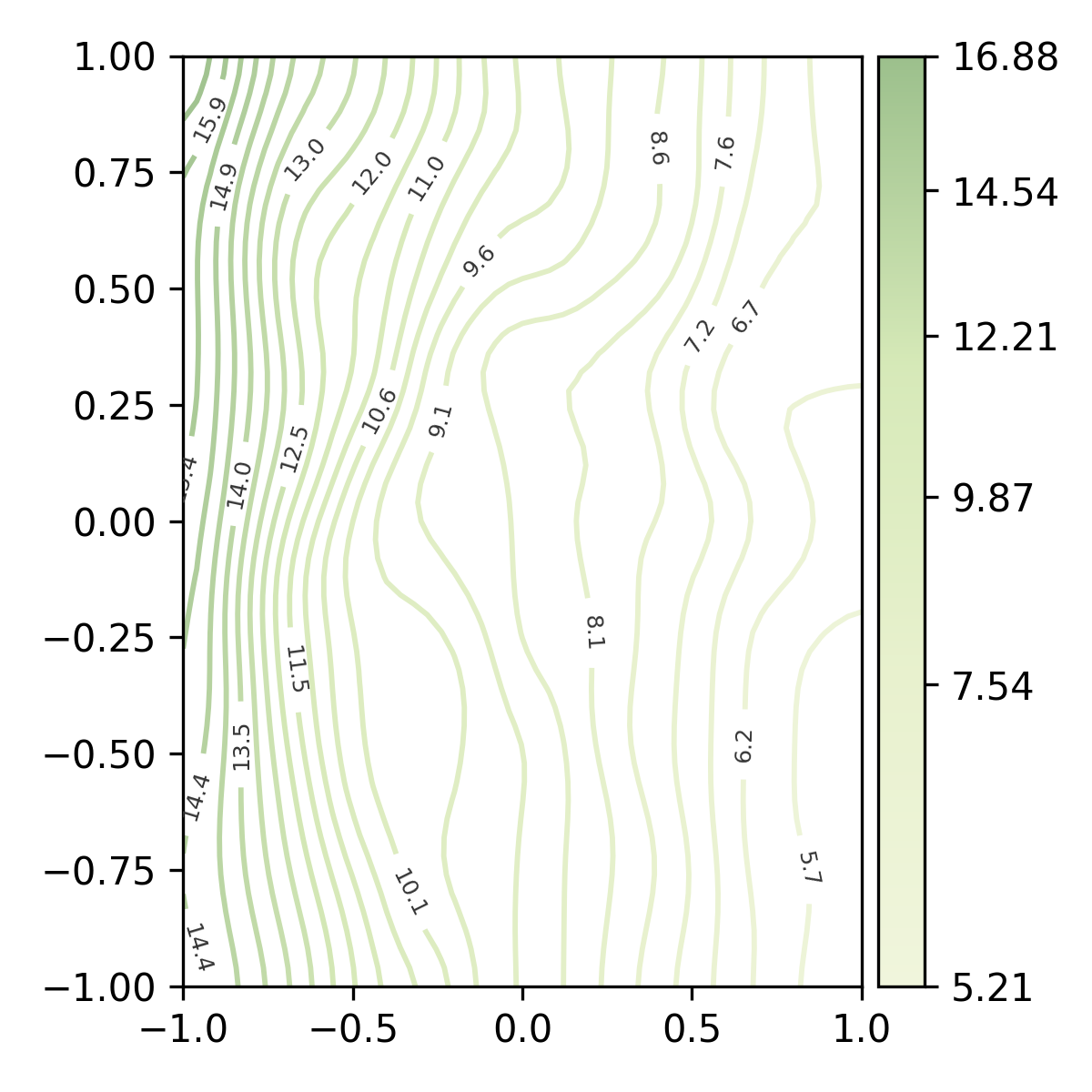}
      \end{minipage}
      \caption{REDDIT}
    \end{subfigure}
    \par\vspace{0.5ex}
    \begin{subfigure}[b]{0.49\linewidth}
      \centering
      \begin{minipage}[c][1.45in][c]{\linewidth}
        \centering
        \includegraphics[width=\linewidth,height=1.45in,keepaspectratio,trim={6bp 6bp 2.4bp 4.8bp},clip]{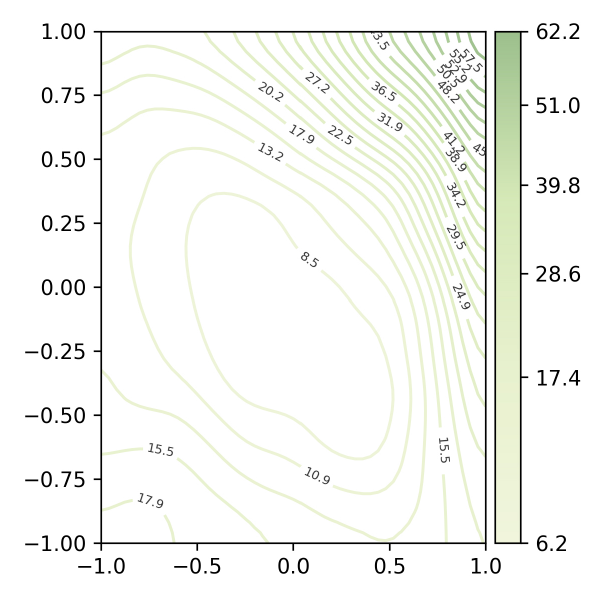}
      \end{minipage}
      \caption{SOCIALEVO}
    \end{subfigure}\hfill
    \begin{subfigure}[b]{0.49\linewidth}
      \centering
      \begin{minipage}[c][1.45in][c]{\linewidth}
        \centering
        \includegraphics[width=\linewidth,height=1.45in,keepaspectratio,trim={6bp 6bp 2.4bp 4.8bp},clip]{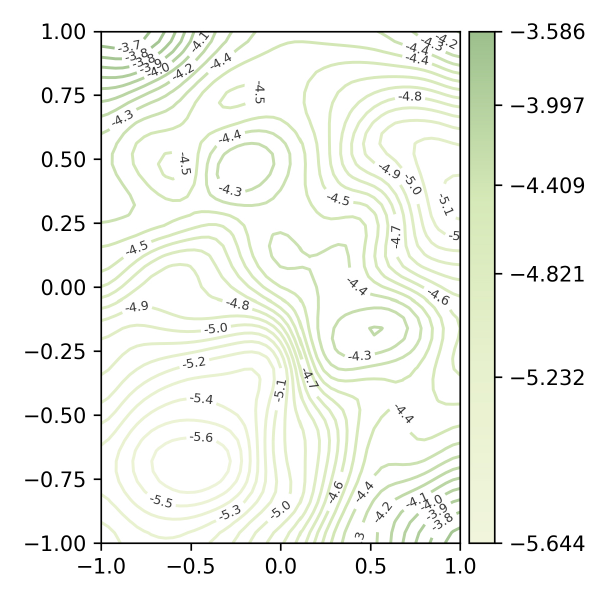}
      \end{minipage}
      \caption{UCI}
    \end{subfigure}
    \caption{2D loss contours of EG-RL on six datasets. Panels (a)--(f): ENRON, LASTFM, MOOC, REDDIT, SOCIALEVO, and UCI. Color bars indicate loss magnitude.}
    \label{fig:loss_landscapes_2d}
    \label{fig:loss_landscapes_all}
  \end{minipage}\hfill
  \begin{minipage}[t]{\columnwidth}
    \vspace{0pt}\centering
    \captionsetup{type=figure}
    \setcounter{subfigure}{0}
    \begin{subfigure}[b]{0.49\linewidth}
      \centering
      \begin{minipage}[c][1.45in][c]{\linewidth}
        \centering
        \includegraphics[width=\linewidth,height=1.45in,keepaspectratio,trim={42bp 2.4bp 12bp 20bp},clip]{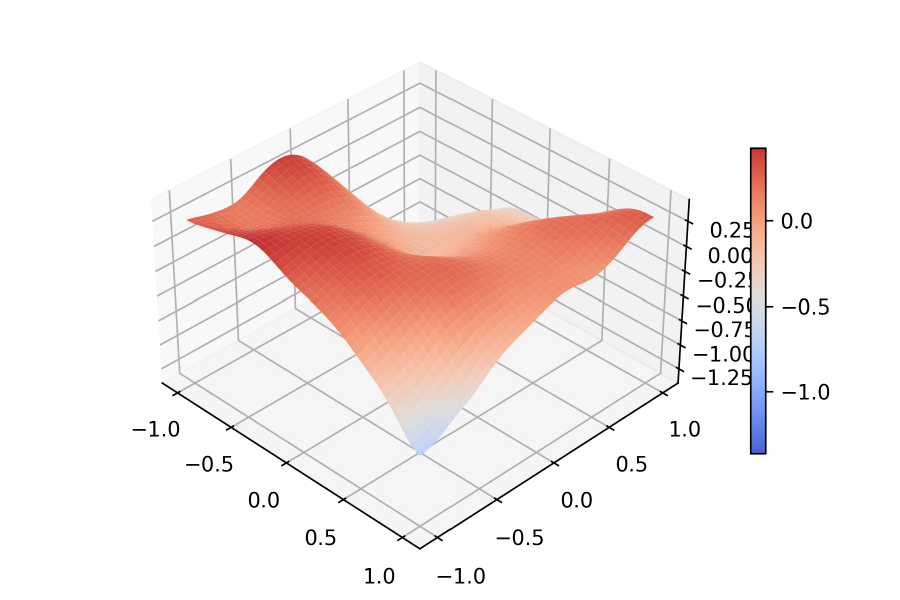}
      \end{minipage}
      \caption{ENRON}
    \end{subfigure}\hfill
    \begin{subfigure}[b]{0.49\linewidth}
      \centering
      \begin{minipage}[c][1.45in][c]{\linewidth}
        \centering
        \includegraphics[width=\linewidth,height=1.45in,keepaspectratio,trim={42bp 2.4bp 12bp 20bp},clip]{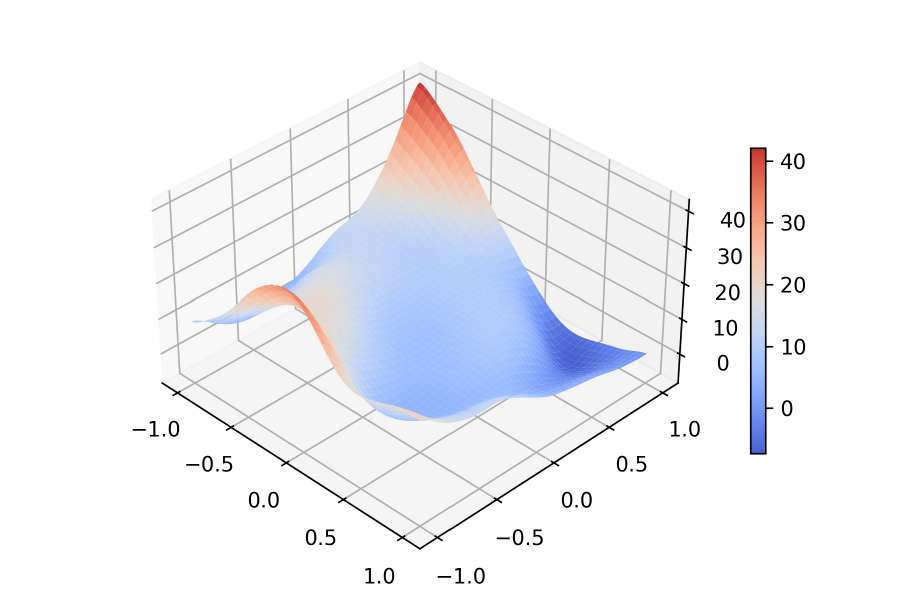}
      \end{minipage}
      \caption{LASTFM}
    \end{subfigure}
    \par\vspace{0.5ex}
    \begin{subfigure}[b]{0.49\linewidth}
      \centering
      \begin{minipage}[c][1.45in][c]{\linewidth}
        \centering
        \includegraphics[width=\linewidth,height=1.45in,keepaspectratio,trim={42bp 2.4bp 12bp 20bp},clip]{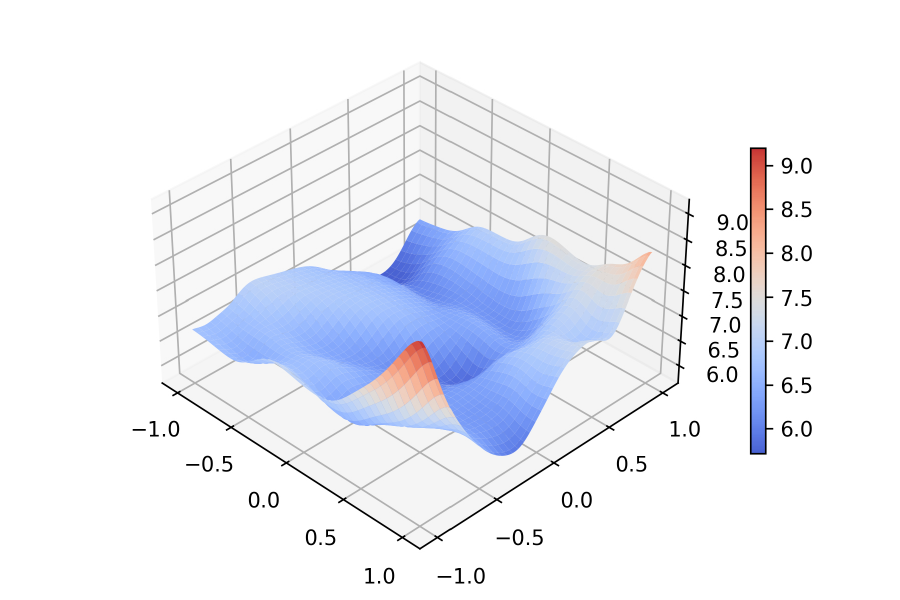}
      \end{minipage}
      \caption{MOOC}
    \end{subfigure}\hfill
    \begin{subfigure}[b]{0.49\linewidth}
      \centering
      \begin{minipage}[c][1.45in][c]{\linewidth}
        \centering
        \includegraphics[width=\linewidth,height=1.45in,keepaspectratio,trim={42bp 2.4bp 12bp 20bp},clip]{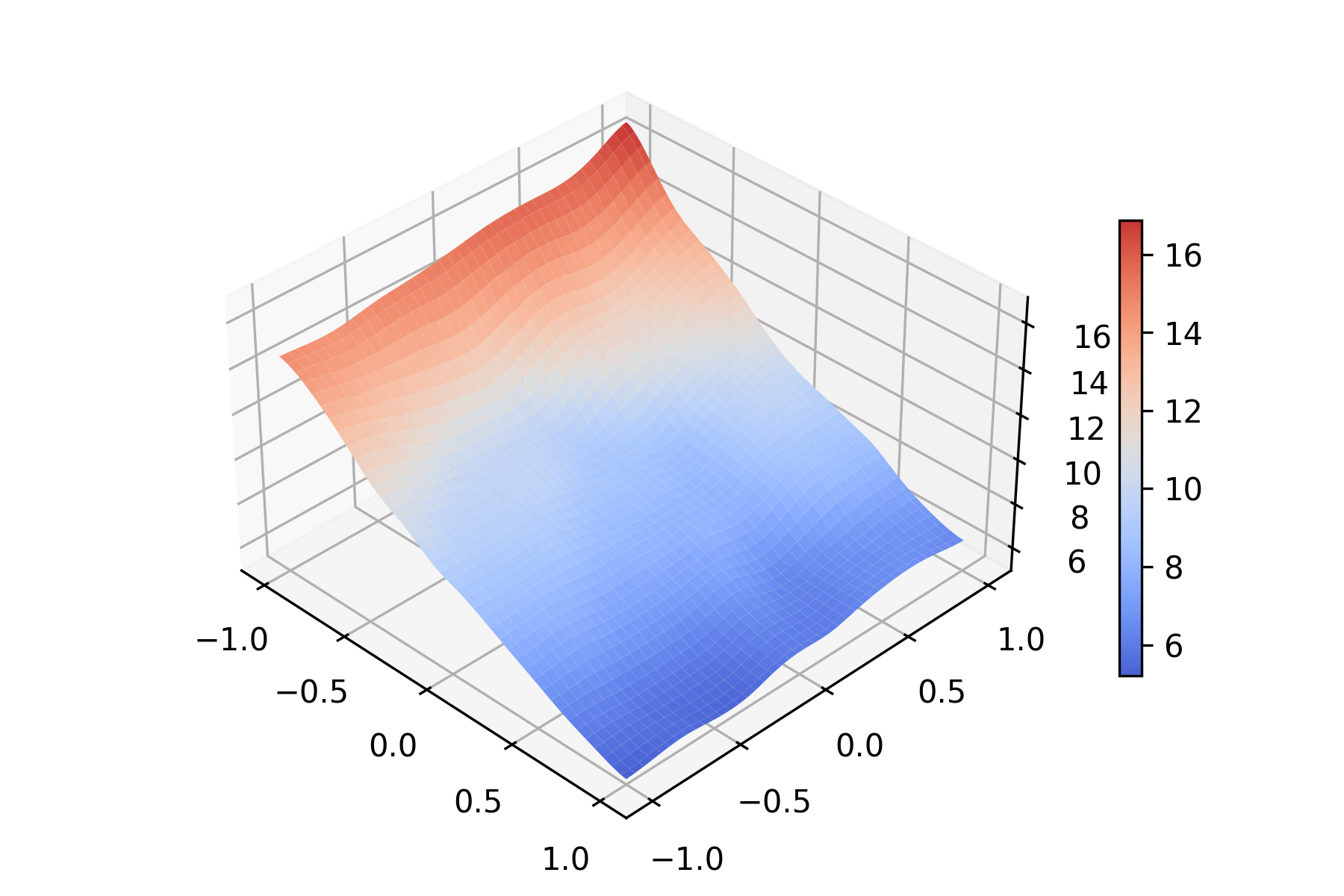}
      \end{minipage}
      \caption{REDDIT}
    \end{subfigure}
    \par\vspace{0.5ex}
    \begin{subfigure}[b]{0.49\linewidth}
      \centering
      \begin{minipage}[c][1.45in][c]{\linewidth}
        \centering
        \includegraphics[width=\linewidth,height=1.45in,keepaspectratio,trim={42bp 2.4bp 12bp 20bp},clip]{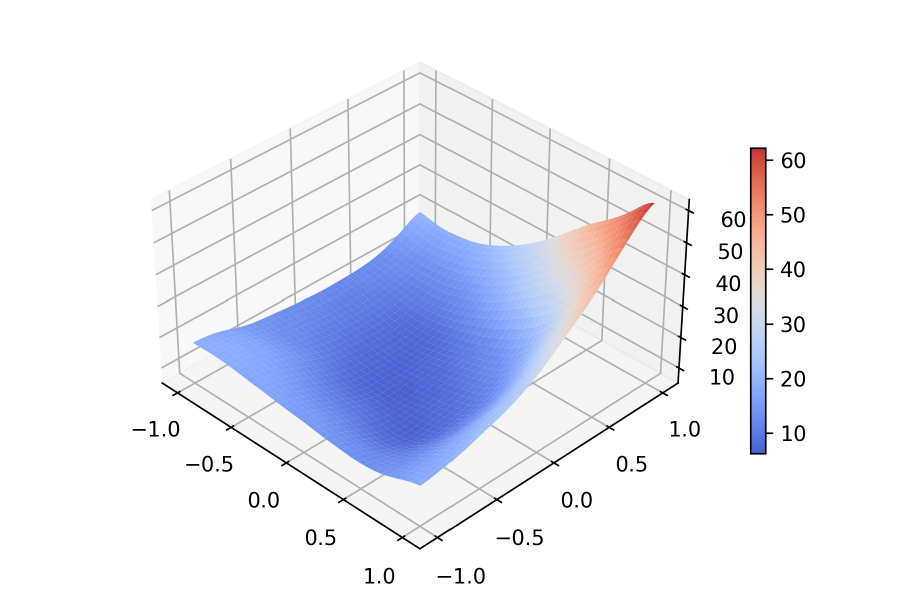}
      \end{minipage}
      \caption{SOCIALEVO}
    \end{subfigure}\hfill
    \begin{subfigure}[b]{0.49\linewidth}
      \centering
      \begin{minipage}[c][1.45in][c]{\linewidth}
        \centering
        \includegraphics[width=\linewidth,height=1.45in,keepaspectratio,trim={42bp 2.4bp 12bp 20bp},clip]{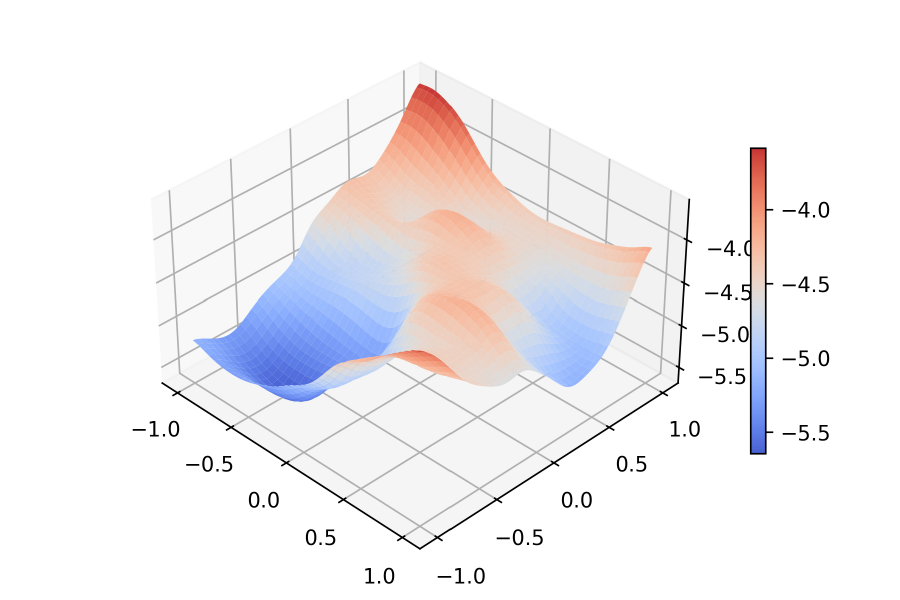}
      \end{minipage}
      \caption{UCI}
    \end{subfigure}
    \caption{3D loss surfaces of EG-RL on the same six datasets, ordered as in Fig.~\ref{fig:loss_landscapes_2d}. Color bars indicate loss magnitude.}
    \label{fig:loss_landscapes_3d}
  \end{minipage}
\end{figure*}

Figure~\ref{fig:loss_landscape_wiki} compares two-parameter-direction
loss slices around converged Wikipedia solutions. EG-RL appears smoother
than DyGFormer and TPNet within the displayed region; this is a qualitative
local observation, not proof of global convexity, better generalization,
or causation of prediction gains. Figures~\ref{fig:loss_landscapes_2d} and~\ref{fig:loss_landscapes_3d}
provide 2D and 3D views on six additional datasets, respectively.

On Wikipedia, the nested contours of EG-RL form an elongated basin
with a comparatively regular progression from its interior toward the
boundary. The corresponding 3D view makes this gradual change in height
visible along both plotted directions. In contrast, DyGFormer's contours
show several pronounced bends, and its surface contains alternating ridges
and valleys. TPNet exhibits a broad central region together with sharper
variation near parts of the boundary.

The additional datasets reveal different local geometries. Social Evolution
has approximately elliptical contours and a broad basin, whereas LastFM
and Reddit show stronger directional slopes. MOOC and UCI contain more
localized undulations, and Enron combines a sloping region with changes
in curvature. Thus, the cross-dataset plots do not simply reproduce the
Wikipedia pattern: they show how the fitted objective varies with the
underlying interaction data.

Each panel has its own color scale, so the visual comparison emphasizes
local shape rather than absolute loss values across models or datasets.

\subsection{Reward Metric Study for the RL Selector}
\label{sec:reward}

\paragraph{Motivation.}
In EG-RL, the snapshot selector is optimized by a scalar reward computed after each RL episode.
The original implementation uses the validation AUC gain as the reward signal.
However, temporal link prediction is commonly reported with both AUC and Average Precision (AP).
To examine the sensitivity of the selector to the choice of reward metric, we additionally evaluate
(i) AP-based reward and (ii) a joint AUC\&\&AP reward.

\paragraph{Reward variants.}
We keep the RL training pipeline unchanged: after an episode, we freeze the encoder, fine-tune the
classifier for one epoch on the augmented graph, and compute the reward on the validation split.
The only difference is how the scalar reward is computed.
Let $G$ denote the original training graph and $G^{aug}$ denote the augmented graph produced by the selector.
We define:

\begin{equation}
r_{\mathrm{AUC}}=\operatorname{clip}\left(\frac{\mathrm{AUC}(G^{aug})-\mathrm{AUC}(G)}{\mathrm{AUC}(G)},[-1,1]\right),
\end{equation}

\begin{equation}
r_{\mathrm{AP}}=\operatorname{clip}\left(\frac{\mathrm{AP}(G^{aug})-\mathrm{AP}(G)}{\mathrm{AP}(G)},[-1,1]\right),
\end{equation}

\begin{equation}
r_{\mathrm{AUC\&\&AP}}=\operatorname{clip}\left(\tfrac{1}{2}r_{\mathrm{AUC}}+\tfrac{1}{2}r_{\mathrm{AP}},[-1,1]\right),
\end{equation}
where $\operatorname{clip}(\cdot,[-1,1])$ truncates values to the interval $[-1,1]$.

\paragraph{Results.}
Table~\ref{tab:reward_metrics} reports the per-dataset scores.
Overall, validation AUC remains the most reliable reward across datasets.
The AP-based reward improves Social Evolution, but generally hurts Enron, UCI, and Flights.
The joint AUC\&\&AP reward acts as a compromise, yet it does not consistently surpass the default AUC reward.
These results indicate that validation AUC provides the most stable signal for snapshot selection in EG-RL.
\begin{table}[!h]
\centering
\caption{Effect of different reward metrics for the RL selector (Test AP / AUC, \%).}
\label{tab:reward_metrics}
\resizebox{\linewidth}{!}{
\begin{tabular}{l|cc|cc|cc|cc}
\toprule
\multirow{2}{*}{Reward} &
\multicolumn{2}{c|}{Wikipedia} &
\multicolumn{2}{c|}{Reddit} &
\multicolumn{2}{c|}{MOOC} &
\multicolumn{2}{c}{LastFM} \\
& AP & AUC & AP & AUC & AP & AUC & AP & AUC \\
\midrule
AUC (default)      & 99.99 & 99.99 & 99.99 & 99.99 & 99.99 & 99.99 & 96.69 & 97.71 \\
AP                 & 99.85 & 99.87 & 99.93 & 99.94 & 99.91 & 99.93 & 96.23 & 97.36 \\
AUC\&\&AP (joint)  & 99.84 & 99.86 & 99.92 & 99.94 & 99.90 & 99.93 & 96.13 & 97.27 \\
\midrule
\multirow{2}{*}{Reward} &
\multicolumn{2}{c|}{Enron} &
\multicolumn{2}{c|}{Social Evo.} &
\multicolumn{2}{c|}{UCI} &
\multicolumn{2}{c}{Flights} \\
& AP & AUC & AP & AUC & AP & AUC & AP & AUC \\
\midrule
AUC (default)      & 92.41 & 93.05 & 94.54 & 95.10 & 96.63 & 96.81 & 98.86 & 98.95 \\
AP                 & 90.98 & 90.15 & 96.98 & 97.30 & 94.88 & 96.15 & 96.72 & 97.55 \\
AUC\&\&AP (joint)  & 90.54 & 90.11 & 96.98 & 97.30 & 94.46 & 96.11 & 96.66 & 97.47 \\
\bottomrule
\end{tabular}}
\end{table}

\subsection{Transductive Results under Alternative Negative Sampling}
\label{sec:neg-sampling}

\paragraph{Motivation.}
Beyond the random negative sampling protocol used in the main results,
we further evaluate EG-RL under two more challenging transductive
settings: historical negatives sampled from previously active nodes,
and inductive negatives sampled from nodes that are inactive at the
prediction time. These protocols change the candidate distribution and
therefore test whether the advantage observed with random negatives
persists under alternative evaluation conditions. We examine both the
average rank across datasets and the margin over the strongest baseline
within each dataset. All margins below are expressed in percentage points.

\paragraph{Historical negative sampling.}
Table~\ref{tab:hist_ns} shows that EG-RL ranks first on all eight datasets
in both AP and AUC, yielding an average rank of 1.00 for each metric.
The strongest competing average ranks are 3.62 in AP and 3.25 in AUC,
both achieved by PINT. The advantage is particularly large on Flights:
EG-RL reaches 97.99\% AP and 98.17\% AUC, exceeding the strongest
baselines by 25.61 and 23.53 points, respectively. On Reddit, its AP of
99.94\% and AUC of 99.95\% improve over PINT by 16.01 and 16.97 points.
LastFM also shows substantial gains of 9.24 points in AP and 13.24 points
in AUC over TPNet. The gains are not uniform across datasets. On UCI,
for example, the AP margin over PINT is 0.51 points, while the AUC margin
is 4.08 points. Reporting both metrics thus reveals differences that a
single aggregate score would obscure.

\paragraph{Inductive negative sampling.}
Table~\ref{tab:inductive_ns} gives EG-RL the best average ranks of 1.38
in AP and 1.25 in AUC. It achieves the highest AP and AUC on seven of
the eight datasets. On MOOC, EG-RL obtains 99.90\% AP and 99.93\% AUC,
exceeding TPNet by 11.72 and 10.86 points. On LastFM, its AP advantage over TPNet is 18.23 points, while its
AUC advantage over EdgeBank is 19.98 points. Flights shows improvements
of 16.15 and 16.53 points over EdgeBank.
Social Evolution provides a contrasting case: EG-RL reaches 96.96\% AP
and 97.31\% AUC, compared with the best scores of 97.68\% and 97.59\%.
The respective gaps are 0.72 and 0.28 points. The strongest competing average ranks are 4.50 in AP for DyGFormer
and 5.38 in AUC for TPNet, indicating a substantial separation in overall
ranking across the eight datasets.

\paragraph{Summary.}
Across the three reported protocols, EG-RL maintains the best average
rank in both AP and AUC: 1.75/1.88 with random negatives, 1.00/1.00
with historical negatives, and 1.38/1.25 with inductive negatives.
The strongest baseline also changes across datasets and protocols;
for example, the Flights historical results favor TGAT in AP and
EdgeBank in AUC, whereas EdgeBank leads the competing methods in
both metrics under inductive sampling. EG-RL's advantage therefore
extends across different competitive reference points. These comparisons
complement the ablations: interpolation supplies explicit transition
cues, and learned retention selects useful augmented inputs. 

\providecommand{\pms}[1]{\,{\scriptsize$\pm$#1}}
\setcounter{dbltopnumber}{1}
\begin{table*}[t]
\centering
\small
\setlength{\tabcolsep}{3pt}
\caption{Transductive results for historical negative sampling. Best result in \textbf{bold}; runner-up \underline{underlined}.}
\label{tab:hist_ns}
\resizebox{\textwidth}{!}{%
\begin{tabular}{@{}llccccccccccccc@{}}
\toprule
\multicolumn{2}{c}{Metric} & JODIE & DyRep & TGAT & TGN & CAWN & EdgeBank & TCL & GraphMixer & NAT & PINT & DyGFormer & TPNet & EG-RL \\
\midrule
\multirow{8}{*}{\rotatebox{90}{\textbf{AP} (\%)}} 
 & Wikipedia   & 83.01\pms{0.66} & 79.93\pms{0.56} & 87.38\pms{0.22} & 86.86\pms{0.33} & 71.21\pms{1.67} & 73.35\pms{0.00} & 89.05\pms{0.39} & 90.90\pms{0.10} & 68.36\pms{4.47} & \underline{91.33}\pms{0.76} & 82.23\pms{2.54} & 81.55\pms{4.10} & \textbf{99.86}\pms{0.02} \\
 & Reddit      & 80.03\pms{0.36} & 79.83\pms{0.31} & 79.55\pms{0.20} & 81.22\pms{0.61} & 80.82\pms{0.45} & 73.59\pms{0.00} & 77.14\pms{0.16} & 78.44\pms{0.18} & 80.58\pms{0.91} & \underline{83.93}\pms{1.11} & 81.57\pms{0.67} & 81.02\pms{1.31} & \textbf{99.94}\pms{0.01} \\
 & MOOC        & 78.94\pms{1.25} & 75.60\pms{1.12} & 82.19\pms{0.62} & 87.06\pms{1.93} & 74.05\pms{0.95} & 60.71\pms{0.00} & 77.06\pms{0.41} & 77.77\pms{0.92} & 85.42\pms{3.26} & 84.80\pms{1.57} & 85.85\pms{0.66} & \underline{92.69}\pms{0.95} & \textbf{99.92}\pms{0.03} \\
 & LastFM      & 74.35\pms{3.81} & 74.92\pms{2.46} & 71.59\pms{0.24} & 76.87\pms{4.64} & 69.86\pms{0.43} & 73.03\pms{0.00} & 59.30\pms{2.31} & 72.47\pms{0.49} & 78.75\pms{1.63} & 84.95\pms{0.43} & 81.57\pms{0.48} & \underline{87.74}\pms{0.50} & \textbf{96.98}\pms{0.05} \\
 & Enron       & 69.85\pms{2.70} & 71.19\pms{2.76} & 64.07\pms{1.05} & 73.91\pms{1.76} & 64.73\pms{0.36} & 76.53\pms{0.00} & 70.66\pms{0.39} & 77.98\pms{0.92} & 72.90\pms{1.38} & \underline{85.41}\pms{0.84} & 75.63\pms{0.73} & 80.79\pms{1.68} & \textbf{90.27}\pms{0.12} \\
 & Social Evo. & 87.44\pms{6.78} & 93.29\pms{0.43} & 95.01\pms{0.44} & 94.45\pms{0.56} & 85.53\pms{0.38} & 80.57\pms{0.00} & 94.74\pms{0.31} & 94.93\pms{0.31} & 91.29\pms{4.89} & 96.73\pms{0.29} & \underline{97.38}\pms{0.14} & 96.80\pms{0.41} & \textbf{98.53}\pms{0.04} \\
 & UCI         & 75.24\pms{5.80} & 55.10\pms{3.14} & 68.27\pms{1.37} & 80.43\pms{2.12} & 65.30\pms{0.43} & 65.50\pms{0.00} & 80.25\pms{2.74} & 84.11\pms{1.35} & 75.38\pms{1.28} & \underline{93.96}\pms{0.20} & 82.17\pms{0.82} & 86.34\pms{0.80} & \textbf{94.47}\pms{0.08} \\
 & Flights     & 66.48\pms{2.59} & 67.61\pms{0.99} & \underline{72.38}\pms{0.18} & 66.70\pms{1.64} & 64.72\pms{0.97} & 70.53\pms{0.00} & 70.68\pms{0.24} & 71.47\pms{0.26} & 64.74\pms{1.83} & 66.82\pms{1.44} & 66.59\pms{0.49} & 69.10\pms{1.27} & \textbf{97.99}\pms{0.04} \\
 \addlinespace[2pt]
 \multicolumn{2}{r}{Avg. Rank} 
 & 9.12 & 9.38 & 7.88 & 6.12 & 11.38 & 10.00 & 8.38 & 6.25 & 8.50 & \underline{3.62} & 5.25 & 4.12 & \textbf{1.00} \\
\midrule
\multirow{8}{*}{\rotatebox{90}{\textbf{AUC} (\%)}} 
 & Wikipedia   & 80.77\pms{0.73} & 77.74\pms{0.33} & 82.87\pms{0.22} & 82.74\pms{0.32} & 67.84\pms{0.64} & 77.27\pms{0.00} & 85.76\pms{0.46} & 87.68\pms{0.17} & 69.32\pms{2.78} & \underline{89.25}\pms{0.49} & 78.80\pms{1.95} & 79.89\pms{2.47} & \textbf{99.87}\pms{0.01} \\
 & Reddit      & 80.52\pms{0.32} & 80.15\pms{0.18} & 79.33\pms{0.16} & 81.11\pms{0.19} & 80.27\pms{0.30} & 78.58\pms{0.00} & 76.49\pms{0.16} & 77.80\pms{0.12} & 79.36\pms{0.31} & \underline{82.98}\pms{0.63} & 80.54\pms{0.29} & 81.87\pms{0.49} & \textbf{99.95}\pms{0.01} \\
 & MOOC        & 82.75\pms{0.83} & 81.06\pms{0.94} & 80.81\pms{0.67} & 88.00\pms{1.80} & 71.57\pms{1.07} & 61.90\pms{0.00} & 72.09\pms{0.56} & 76.68\pms{1.40} & 84.93\pms{2.89} & 87.44\pms{1.74} & 87.04\pms{0.35} & \underline{93.45}\pms{0.67} & \textbf{99.94}\pms{0.02} \\
 & LastFM      & 75.22\pms{2.36} & 74.65\pms{1.98} & 64.27\pms{0.26} & 77.97\pms{3.04} & 67.88\pms{0.24} & 78.09\pms{0.00} & 47.24\pms{3.13} & 64.21\pms{0.73} & 75.89\pms{2.21} & 81.89\pms{0.96} & 78.78\pms{0.35} & \underline{84.64}\pms{0.45} & \textbf{97.88}\pms{0.04} \\
 & Enron       & 75.39\pms{2.37} & 74.69\pms{3.55} & 61.85\pms{1.43} & 77.09\pms{2.22} & 65.10\pms{0.34} & 79.59\pms{0.00} & 67.95\pms{0.88} & 75.27\pms{1.14} & 73.22\pms{2.18} & \underline{83.80}\pms{0.68} & 76.55\pms{0.52} & 81.16\pms{1.28} & \textbf{89.92}\pms{0.10} \\
 & Social Evo. & 90.06\pms{3.15} & 93.12\pms{0.34} & 93.08\pms{0.59} & 94.71\pms{0.53} & 87.43\pms{0.15} & 85.81\pms{0.00} & 93.44\pms{0.68} & 94.39\pms{0.31} & 91.87\pms{4.52} & 96.82\pms{0.24} & \underline{97.28}\pms{0.07} & 97.22\pms{0.30} & \textbf{98.75}\pms{0.03} \\
 & UCI         & 78.64\pms{3.50} & 57.91\pms{3.12} & 58.89\pms{1.57} & 77.25\pms{2.68} & 57.86\pms{0.15} & 69.56\pms{0.00} & 72.25\pms{3.46} & 77.54\pms{2.02} & 71.52\pms{1.61} & \underline{92.05}\pms{0.36} & 76.97\pms{0.24} & 80.42\pms{0.64} & \textbf{96.13}\pms{0.07} \\
 & Flights     & 68.97\pms{1.87} & 69.43\pms{0.90} & 72.20\pms{0.16} & 68.39\pms{0.95} & 66.11\pms{0.71} & \underline{74.64}\pms{0.00} & 70.57\pms{0.18} & 70.37\pms{0.23} & 67.11\pms{2.16} & 69.52\pms{0.90} & 68.09\pms{0.43} & 71.82\pms{0.82} & \textbf{98.17}\pms{0.04} \\
 \addlinespace[2pt]
 \multicolumn{2}{r}{Avg. Rank} 
 & 7.38 & 9.00 & 8.88 & 5.62 & 11.50 & 8.62 & 9.00 & 7.75 & 9.38 & \underline{3.25} & 6.12 & 3.50 & \textbf{1.00} \\
\bottomrule
\end{tabular}
}
\end{table*}
\providecommand{\pms}[1]{\,{\scriptsize$\pm$#1}}
\begin{table*}[t]
\centering
\small
\setlength{\tabcolsep}{3pt}
\caption{Transductive results for inductive negative sampling. Best result in \textbf{bold}; runner-up \underline{underlined}.}
\label{tab:inductive_ns}
\resizebox{\textwidth}{!}{%
\begin{tabular}{@{}llccccccccccccc@{}}
\toprule
\multicolumn{2}{c}{Metric} & JODIE & DyRep & TGAT & TGN & CAWN & EdgeBank & TCL & GraphMixer & NAT & PINT & DyGFormer & TPNet & EG-RL \\
\midrule
\multirow{8}{*}{\rotatebox{90}{\textbf{AP} (\%)}} 
 & Wikipedia   & 75.65\pms{0.79} & 70.21\pms{1.58} & 87.00\pms{0.16} & 85.62\pms{0.44} & 74.06\pms{2.62} & 80.63\pms{0.00} & 86.76\pms{0.72} & \underline{88.59}\pms{0.17} & 53.85\pms{6.25} & 82.39\pms{2.70} & 78.29\pms{5.38} & 79.35\pms{5.52} & \textbf{99.84}\pms{0.02} \\
 & Reddit      & 86.98\pms{0.16} & 86.30\pms{0.26} & 89.59\pms{0.24} & 88.10\pms{0.24} & \underline{91.67}\pms{0.24} & 85.48\pms{0.00} & 87.45\pms{0.29} & 85.26\pms{0.11} & 81.71\pms{1.03} & 87.59\pms{0.65} & 91.11\pms{0.40} & 88.19\pms{0.33} & \textbf{99.92}\pms{0.01} \\
 & MOOC        & 65.23\pms{2.19} & 61.66\pms{0.95} & 75.95\pms{0.64} & 77.50\pms{2.91} & 73.51\pms{0.94} & 49.43\pms{0.00} & 74.65\pms{0.54} & 74.27\pms{0.92} & 77.28\pms{1.94} & 75.70\pms{1.43} & 81.24\pms{0.69} & \underline{88.18}\pms{0.97} & \textbf{99.90}\pms{0.02} \\
 & LastFM      & 62.67\pms{4.49} & 64.41\pms{2.70} & 71.13\pms{0.17} & 65.95\pms{5.98} & 67.48\pms{0.77} & 75.49\pms{0.00} & 58.21\pms{0.89} & 68.12\pms{0.33} & 64.02\pms{1.48} & 72.24\pms{0.92} & 73.97\pms{0.50} & \underline{77.99}\pms{1.30} & \textbf{96.22}\pms{0.05} \\
 & Enron       & 68.96\pms{0.98} & 67.79\pms{1.53} & 63.94\pms{1.36} & 70.89\pms{2.72} & 75.15\pms{0.58} & 73.89\pms{0.00} & 71.29\pms{0.32} & 75.01\pms{0.79} & 70.34\pms{1.73} & \underline{80.23}\pms{1.37} & 77.41\pms{0.89} & 75.36\pms{1.81} & \textbf{89.55}\pms{0.10} \\
 & Social Evo. & 89.82\pms{4.11} & 93.28\pms{0.48} & 94.84\pms{0.44} & 95.13\pms{0.56} & 88.32\pms{0.27} & 83.69\pms{0.00} & 94.90\pms{0.36} & 94.72\pms{0.33} & 92.12\pms{4.53} & 97.15\pms{0.25} & \textbf{97.68}\pms{0.10} & \underline{97.35}\pms{0.32} & 96.96\pms{0.05} \\
 & UCI         & 65.99\pms{1.40} & 54.79\pms{1.76} & 68.67\pms{0.84} & 70.94\pms{0.71} & 64.61\pms{0.48} & 57.43\pms{0.00} & 76.01\pms{1.11} & 80.10\pms{0.51} & 57.55\pms{0.81} & \underline{83.61}\pms{0.23} & 72.25\pms{1.71} & 77.26\pms{1.57} & \textbf{94.23}\pms{0.08} \\
 & Flights     & 69.07\pms{4.02} & 70.57\pms{1.82} & 75.48\pms{0.26} & 71.09\pms{2.72} & 69.18\pms{1.52} & \underline{81.08}\pms{0.00} & 74.62\pms{0.18} & 74.87\pms{0.21} & 59.16\pms{1.48} & 61.99\pms{1.70} & 70.92\pms{1.78} & 64.78\pms{1.50} & \textbf{97.23}\pms{0.04} \\
 \addlinespace[2pt]
 \multicolumn{2}{r}{Avg. Rank} 
 & 10.38 & 10.75 & 6.25 & 6.38 & 8.38 & 8.50 & 7.12 & 6.38 & 10.75 & 5.50 & \underline{4.50} & 4.75 & \textbf{1.38} \\
\midrule
\multirow{8}{*}{\rotatebox{90}{\textbf{AUC} (\%)}} 
 & Wikipedia   & 70.96\pms{0.78} & 67.36\pms{0.96} & 81.93\pms{0.22} & 80.97\pms{0.31} & 70.95\pms{0.95} & 81.73\pms{0.00} & 82.19\pms{0.48} & \underline{84.28}\pms{0.30} & 52.11\pms{5.83} & 77.03\pms{1.93} & 75.09\pms{3.70} & 75.36\pms{3.41} & \textbf{99.86}\pms{0.02} \\
 & Reddit      & 83.51\pms{0.15} & 82.90\pms{0.31} & 87.13\pms{0.20} & 84.56\pms{0.24} & \underline{88.04}\pms{0.29} & 85.93\pms{0.00} & 84.67\pms{0.29} & 82.21\pms{0.13} & 76.01\pms{1.04} & 81.02\pms{0.68} & 86.23\pms{0.51} & 81.64\pms{0.42} & \textbf{99.94}\pms{0.01} \\
 & MOOC        & 66.63\pms{2.30} & 63.26\pms{1.01} & 73.18\pms{0.33} & 77.44\pms{2.86} & 70.32\pms{1.43} & 48.18\pms{0.00} & 70.36\pms{0.37} & 72.45\pms{0.72} & 75.56\pms{2.39} & 77.00\pms{2.24} & 80.76\pms{0.76} & \underline{89.07}\pms{0.63} & \textbf{99.93}\pms{0.02} \\
 & LastFM      & 61.32\pms{3.49} & 62.15\pms{2.12} & 63.99\pms{0.21} & 65.46\pms{4.27} & 67.92\pms{0.44} & \underline{77.37}\pms{0.00} & 46.93\pms{2.59} & 60.22\pms{0.32} & 58.54\pms{2.37} & 69.51\pms{0.61} & 69.25\pms{0.36} & 72.48\pms{0.90} & \textbf{97.35}\pms{0.04} \\
 & Enron       & 70.92\pms{1.05} & 68.73\pms{1.34} & 60.45\pms{2.12} & 71.34\pms{2.46} & 75.17\pms{0.50} & 75.00\pms{0.00} & 67.64\pms{0.86} & 71.53\pms{0.85} & 69.62\pms{1.12} & \underline{78.91}\pms{0.96} & 74.07\pms{0.64} & 75.44\pms{1.38} & \textbf{89.42}\pms{0.09} \\
 & Social Evo. & 90.01\pms{3.19} & 93.07\pms{0.38} & 92.94\pms{0.61} & 95.24\pms{0.56} & 89.93\pms{0.15} & 87.88\pms{0.00} & 93.44\pms{0.72} & 94.22\pms{0.32} & 92.53\pms{4.20} & 97.10\pms{0.22} & \underline{97.51}\pms{0.06} & \textbf{97.59}\pms{0.25} & 97.31\pms{0.04} \\
 & UCI         & 64.14\pms{1.26} & 54.25\pms{2.01} & 60.80\pms{1.01} & 64.11\pms{1.04} & 58.06\pms{0.26} & 58.03\pms{0.00} & 70.05\pms{1.86} & 74.59\pms{0.74} & 49.73\pms{1.20} & \underline{79.42}\pms{0.31} & 65.96\pms{1.18} & 70.85\pms{0.96} & \textbf{95.57}\pms{0.05} \\
 & Flights     & 69.99\pms{3.10} & 71.13\pms{1.55} & 73.47\pms{0.18} & 71.63\pms{1.72} & 69.70\pms{0.75} & \underline{81.10}\pms{0.00} & 72.54\pms{0.19} & 72.21\pms{0.21} & 59.73\pms{1.61} & 59.61\pms{1.65} & 69.53\pms{1.17} & 64.21\pms{1.30} & \textbf{97.63}\pms{0.03} \\
 \addlinespace[2pt]
 \multicolumn{2}{r}{Avg. Rank} 
 & 9.25 & 10.00 & 7.00 & 6.38 & 8.00 & 7.00 & 7.38 & 6.50 & 11.12 & 6.12 & 5.62 & \underline{5.38} & \textbf{1.25} \\
\bottomrule
\end{tabular}
}
\end{table*}

\input{learning_dynamics}

\subsection{Case Study}
\label{subsec:case_study}
\input{case_study}

\input{budget_sensitivity}

\section{Conclusion}
\label{sec:conclusion}
We present EG-RL, a temporal link-prediction framework that enriches binary
edge observations with linearly interpolated gradients and uses an
actor--critic module to select informative snapshots. A token--channel
Mixer encoder yields competitive AUC and average precision, while the
original ablations examine the roles of interpolation and adaptive
selection. Local loss-landscape views provide qualitative
optimization evidence, not proof of global convexity or generalization.
The evaluation on eight benchmarks yields the best average AP and AUC
ranks across the reported protocols. Historical and inductive
negative-sampling results further demonstrate effectiveness with
challenging candidate sets. Ablations confirm the contributions of
interpolation and adaptive selection. On MOOC, the matched comparison
achieves a 29.5\% reduction in final validation BCE at the same update
budget and reaches a comparable loss earlier. The edge-evolution case
study illustrates how increasing and decreasing weights encode observed
appearance and disappearance phases.
Together, these findings highlight the value of combining explicit
temporal transition cues with learned retention in a lightweight encoder.
Future work will investigate more flexible interpolation profiles and
lower-cost selection strategies, and evaluate larger temporal graphs
with different event densities and observation granularities.

\bibliographystyle{IEEEtran}
\bibliography{references}

\end{document}

%% file: learning_dynamics.tex
\subsection{Learning Dynamics}
\label{subsec:learning_dynamics}
We compare EG-RL with its complete edge-gradient augmentation, including
the RL-based retention mechanism, against an original-graph control trained
without augmentation. Both models start from identical parameter
initializations and use the same training examples, frozen negative
samples, batch size of 600, and 20 complete epochs. These learning curves
assess optimization, whereas the following case study examines the
representation of observed edge-state changes. The original-graph control samples
neighborhoods directly from the original binary history,
before top-$k$ selection.

\begin{figure}[!t]
\centering
\includegraphics[width=\linewidth]{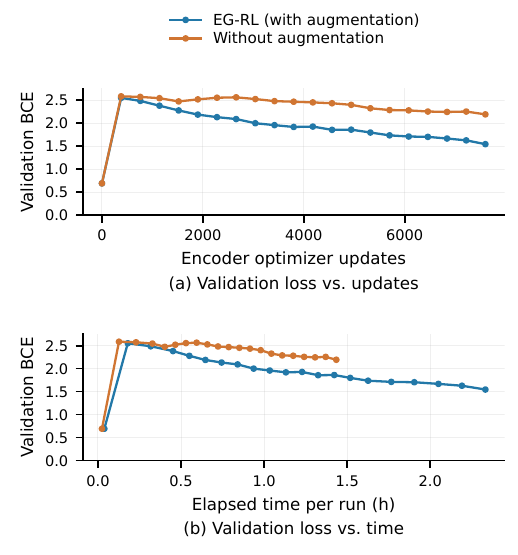}
\caption{Validation-loss dynamics on MOOC with and without edge-gradient
augmentation. The complete augmentation includes RL-based selection.
The same frozen validation set is evaluated before training and after
each epoch. (a) BCE against encoder optimizer updates.
(b) The same BCE against actual elapsed time for each run.}
\label{fig:mooc_learning_dynamics}
\end{figure}

\begin{figure}[!t]
\centering
\includegraphics[width=\linewidth]{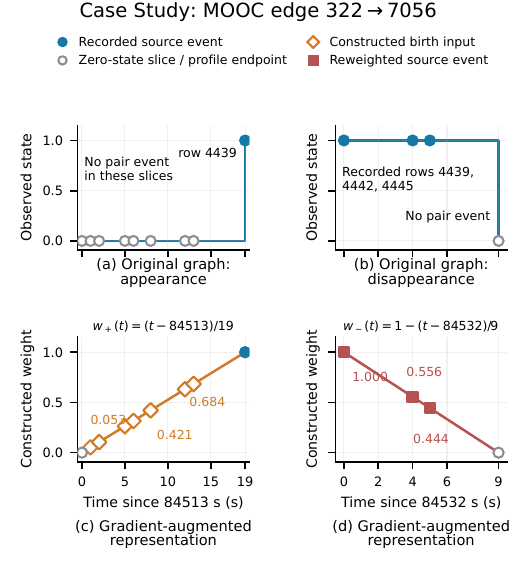}
\caption{Case Study: observed edge evolution on MOOC for learner 322 and
activity 7056. (a, b) Original binary presence at the actual observation
slices. (c) Constructed increasing weights for an observed $0\!\to\!1$
transition. Hollow diamonds are virtual inputs, not recorded interactions.
(d) Decreasing weights applied to the existing positive event rows for
the subsequent $1\!\to\!0$ transition. The zero endpoint is a profile
boundary, not an inserted event. The profiles are complete only at their
closing timestamps and usable only for strictly later queries.}
\label{fig:mooc_edge_evolution}
\end{figure}

Figure~\ref{fig:mooc_learning_dynamics} reports mean binary cross-entropy
(BCE) over all labels of 61,762 fixed validation positives and their
paired random negatives. This is a learning-dynamics diagnostic, not a
historical-negative test result; no test labels are used. Following the
initial loss increase, EG-RL exhibits faster validation-loss reduction.
At 3,800 encoder updates, validation BCE is 1.920 with augmentation
versus 2.469 without it. After five epochs, the augmented model reaches
2.189 in approximately 39 minutes, whereas the original-graph model records
a similar value of 2.194 after 20 epochs and approximately 86 minutes.
At epoch 20, validation BCE is
1.547 with augmentation versus 2.194 without it.

Overall, RL-guided edge-gradient augmentation yields lower validation
BCE at all 20 epoch-end evaluations. The reduction is 22.2\% at 3,800
updates and 29.5\% after 20 epochs (7,600 updates), demonstrating better
validation-loss performance at a fixed update budget. Reaching a comparable
BCE of approximately 2.19 in 39 rather than 86 minutes further shows that
faster loss reduction can offset the augmentation overhead at this observed
loss level.

Figure~\ref{fig:mooc_edge_evolution} complements these empirical findings
by showing how augmentation encodes the direction and progress of
observed edge-state transitions.

%% file: case_study.tex
We examine how edge-gradient augmentation represents observed edge
evolution, rather than using the case study as another performance
comparison. MOOC records time-stamped learner--activity
interactions~\cite{kumar2019predicting}. At each original global timestamp,
a pair has state one if an interaction is recorded and zero otherwise.
A zero therefore denotes no recorded interaction in that slice, not a
verified termination of an underlying relationship.

Figure~\ref{fig:mooc_edge_evolution} follows learner 322 and activity 7056.
We select the earliest completed appearance--disappearance cycle with a
preceding observed interaction and at least two interior observation
slices in each transition interval. This source-only rule yields the
first of 236 eligible cycles; no prediction score or outcome determines
the choice. All timestamps, event rows, and constructed weights are
checked against the original event stream and the actual augmentation
implementation.

\paragraph{Observed appearance, $0\!\to\!1$.}
After an interaction at 84512 s (source row 4426), the pair is absent
from the slice at 84513 s and remains unobserved in the intervening
slices until 84532 s, when row 4439 records another interaction.
The original representation shows a binary jump
(Fig.~\ref{fig:mooc_edge_evolution}(a)).
Over this 19-second interval, the constructed profile is
$w_+(t)=(t-84513)/19$. The seven actual interior slice times receive
fractional weights; for example, 84514, 84521, and 84526 s receive
0.0526, 0.4211, and 0.6842, respectively
(Fig.~\ref{fig:mooc_edge_evolution}(c)).
These virtual inputs reuse the feature row of the observed endpoint,
4439. They explicitly encode increasing transition phase but are not
additional learner actions.

\paragraph{Observed disappearance, $1\!\to\!0$.}
The pair is observed at 84532, 84536, and 84537 s (rows 4439, 4442,
and 4445), but not at the next original slice, 84541 s.
The binary representation stays at one across those three occupied
slices before returning to zero
(Fig.~\ref{fig:mooc_edge_evolution}(b)).
The completed nine-second profile
$w_-(t)=1-(t-84532)/9$ assigns weights 1, 0.5556, and 0.4444 to these
same genuine event rows
(Fig.~\ref{fig:mooc_edge_evolution}(d)).
Thus, the implementation represents a decreasing phase by reweighting
existing observations, without fabricating disappearance events.

\paragraph{Relation to learning dynamics.}
These phase cues offer a possible representation-level explanation for
the faster post-initial validation-loss reduction in
Fig.~\ref{fig:mooc_learning_dynamics}, not proof of optimization speedup.
Input-weight differences are not training losses. Efficiency is assessed
by updates and measured time to comparable validation loss, including
augmentation overhead, rather than by profile smoothness alone.

%% file: budget_sensitivity.tex
\subsection{Candidate Budget Sensitivity}
\label{subsec:budget_sensitivity}
We vary the policy examination cap $B$ over 64, 128, 256, and 512 in an
independent Enron/UCI study. Chronological splits and historical negatives
(with random fallback) are fixed. The 4,206/5,932 test positives involve nodes
excluded from positive training, with one negative per positive.
Every cap trains for 20 epochs (batch size 600; seeds 0--2), with initial
weights and training-negative draws matched within each seed. Uniform
training negatives use the full node-ID namespace. We use PyTorch 2.7,
causal gradient inputs, one-hot node encodings, and original edge features.
Encoder Adam uses learning rate $10^{-4}$ and weight decay $10^{-6}$;
actor and critic use $3\!\times\!10^{-4}$. Rewards are clipped relative
validation-AUC gains from epoch one. Caps apply per batch during training,
validation, and testing; unexamined candidates retain constructed weights
and recent neighborhoods remain fixed. Validation AP selects matching
encoder/policy checkpoints before one test pass. AP/AUC equally weight
frozen 200-link batches, including the final partial batch.

\begin{figure}[!t]
\centering
\includegraphics[width=\linewidth]{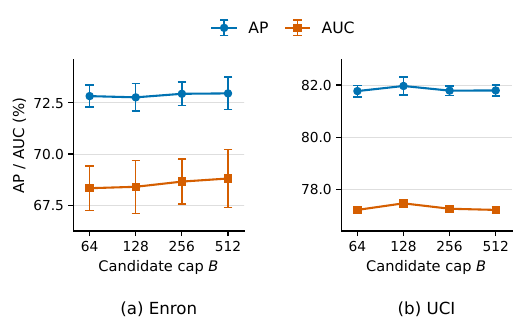}
\caption{Candidate-budget sensitivity under the additional held-out protocol.
Points show test AP/AUC of validation-selected checkpoints (mean and sample
standard deviation, $n=3$). Retention remains learned.}
\label{fig:budget_sensitivity}
\end{figure}

Figure~\ref{fig:budget_sensitivity} shows modest budget sensitivity.
Enron mean AP/AUC span 72.77--72.96\% and 68.33--68.81\%,
while UCI spans 81.77--81.97\% and 77.20--77.46\%.
Enron peaks at $B=512$, whereas UCI peaks at $B=128$
($81.97\!\pm\!0.34\%$ AP;
$77.46\!\pm\!0.16\%$ AUC).
Increasing the cap thus yields dataset-dependent returns. Moderate caps
achieve comparable observed scores with fewer candidate examinations;
the preferred budget should be selected on validation data.

